%% file: main.tex
\documentclass{article}
\usepackage{iclr2027_conference,times}

\input{math_commands.tex}

\usepackage{amsmath,amssymb,amsthm}
\usepackage[table,dvipsnames]{xcolor}
\usepackage{hyperref}
\hypersetup{
    colorlinks=true,
    linkcolor=NavyBlue,
    citecolor=NavyBlue,
    filecolor=NavyBlue,
    urlcolor=NavyBlue
}
\usepackage{url}
\usepackage{marvosym}
\usepackage{graphicx}
\usepackage{wrapfig}
\usepackage{array}
\usepackage{booktabs}
\usepackage{multirow}
\usepackage{longtable}
\usepackage{subcaption}
\usepackage{algorithm}
\usepackage{algpseudocode}
\usepackage{enumitem}
\usepackage{afterpage}
\usepackage[capitalize,noabbrev]{cleveref}

\graphicspath{{figures/}}

\input{macros.tex}

\title{Ground Your Steps: Optimizing Sampling Schedules in Flow Matching Robot Policies}
 
\author{Licheng Yang$^{1,2,3}$, Yonghao He$^{2,\dagger}$, Lingfeng Qian$^{2}$, Xin Shi$^{2}$, Wei Sui$^{2,\text{\Letter}}$, Hu Su$^{1,\text{\Letter}}$ \\[3pt]
\normalfont $^{1}$Institute of Automation, Chinese Academy of Sciences \quad $^{2}$D-Robotics \\[2pt]
\normalfont $^{3}$School of Artificial Intelligence, University of Chinese Academy of Sciences \\[2pt]
\normalfont \texttt{\{yanglicheng2026, hu.su\}@ia.ac.cn, wei.sui@d-robotics.cc}
}

\iclrfinalcopy
 
\begin{document}
 
\maketitle
\begingroup
\renewcommand\thefootnote{}
\begin{NoHyper}
\footnotetext{\hspace{-1.8em}$^{\dagger}$Project lead \quad $^{\text{\Letter}}$Corresponding authors}
\end{NoHyper}
\endgroup
 
\begin{abstract}
Flow matching robot policies generate action chunks by integrating a
learned velocity field at inference.
However, under a fixed budget of neural function evaluations (NFEs),
the integration result depends on the sampling schedule.
This motivates optimizing the schedule to bring the generated action
distribution closer to the demonstrations and improve task success.
We propose \methodlong{} (\method{}), a training-free method that optimizes
the sampling schedule by matching the generated and demonstrated chunk
distributions at each observation.
\method{} applies truncated fractional differencing along each arm
joint's action sequence and compares two types of pairwise differences
between action chunks at each observation: differences of a generated
chunk against a demonstration, and differences of two independent
generated chunks against each other.
After fractional differencing, these two pairwise-difference distributions
coincide exactly when the
distributions of generated and demonstrated action chunks match.
\method{} minimizes the energy distance between these distributions
to obtain an improved schedule for each task and NFE budget.
Compared with the uniform grid, \method{} brings the
generated chunk distribution closer to
the demonstrated distribution and raises the
success rate through the sampling schedule alone.
On \robotwin{} benchmarks, \method{} improves success rates at nearly every budget,
with gains of $2.8$ percentage points for both \pifive{} and \groot{}.
The gains also transfer to three real-world manipulation tasks,
reaching up to $3.3$ percentage points.
Code is available at \url{https://github.com/D-Robotics-AI-Lab/GroundYourSteps}.
\end{abstract}
 
\input{sections/1_intro}
\input{sections/2_related}
\input{sections/3_background}
\input{sections/4_method}
\input{sections/5_experiments}
\input{sections/6_conclusion}

\bibliography{refs}
\bibliographystyle{iclr2027_conference}
 
\appendix
 
\crefalias{section}{appendix}
\crefalias{subsection}{appendix}

\input{sections/appendix}

\end{document}

%% file: math_commands.tex
\usepackage{amsmath,amsfonts,bm}

\def\eqref#1{equation~\ref{#1}}

\def\1{\bm{1}}

\DeclareMathAlphabet{\mathsfit}{\encodingdefault}{\sfdefault}{m}{sl}
\SetMathAlphabet{\mathsfit}{bold}{\encodingdefault}{\sfdefault}{bx}{n}

\newcommand{\E}{\mathbb{E}}

\newcommand{\R}{\mathbb{R}}

\newcommand{\Cov}{\mathrm{Cov}}

%% file: macros.tex
\newcommand{\method}{GYS}
\newcommand{\methodlong}{Ground Your Steps}

\newcommand{\sched}{\boldsymbol{\tau}}
\newcommand{\demos}{q}
\newcommand{\gen}{p}
\newcommand{\grids}{\mathcal{T}}
\newcommand{\feas}{\mathcal{H}_B}

\newcommand{\obsrep}{\xi}

\newcommand{\jcross}{c}
\newcommand{\jself}{s}

\newcommand{\Dh}{D_{h}}

\newcommand{\Law}{\operatorname{Law}}
\newcommand{\crosslaw}{C}
\newcommand{\selflaw}{S}
\newcommand{\horizonop}{\Phi}
\newcommand{\bandord}{\nu}
\newcommand{\energy}{\widehat{\mathcal{E}}}
\newcommand{\obj}{L}
\newcommand{\qstar}{\nu^{\star}}

\newcommand{\pap}{PaP}
\newcommand{\sbc}{SbC}
\newcommand{\sac}{SaC}

\newcommand{\groot}{GR00T~N1.7}
\newcommand{\pifive}{$\pi_{0.5}$}
\newcommand{\robotwin}{RoboTwin~2.0}

\theoremstyle{plain}
\newtheorem{proposition}{Proposition}

\theoremstyle{definition}
\newtheorem{assumption}{Assumption}
\crefname{assumption}{Assumption}{Assumptions}
\crefname{proposition}{Proposition}{Propositions}
\crefname{lemma}{Lemma}{Lemmas}

%% file: sections/1_intro.tex
\section{Introduction}
\label{sec:intro}

\afterpage{%
\begin{figure}[htb]
\centering
\includegraphics[width=\linewidth]{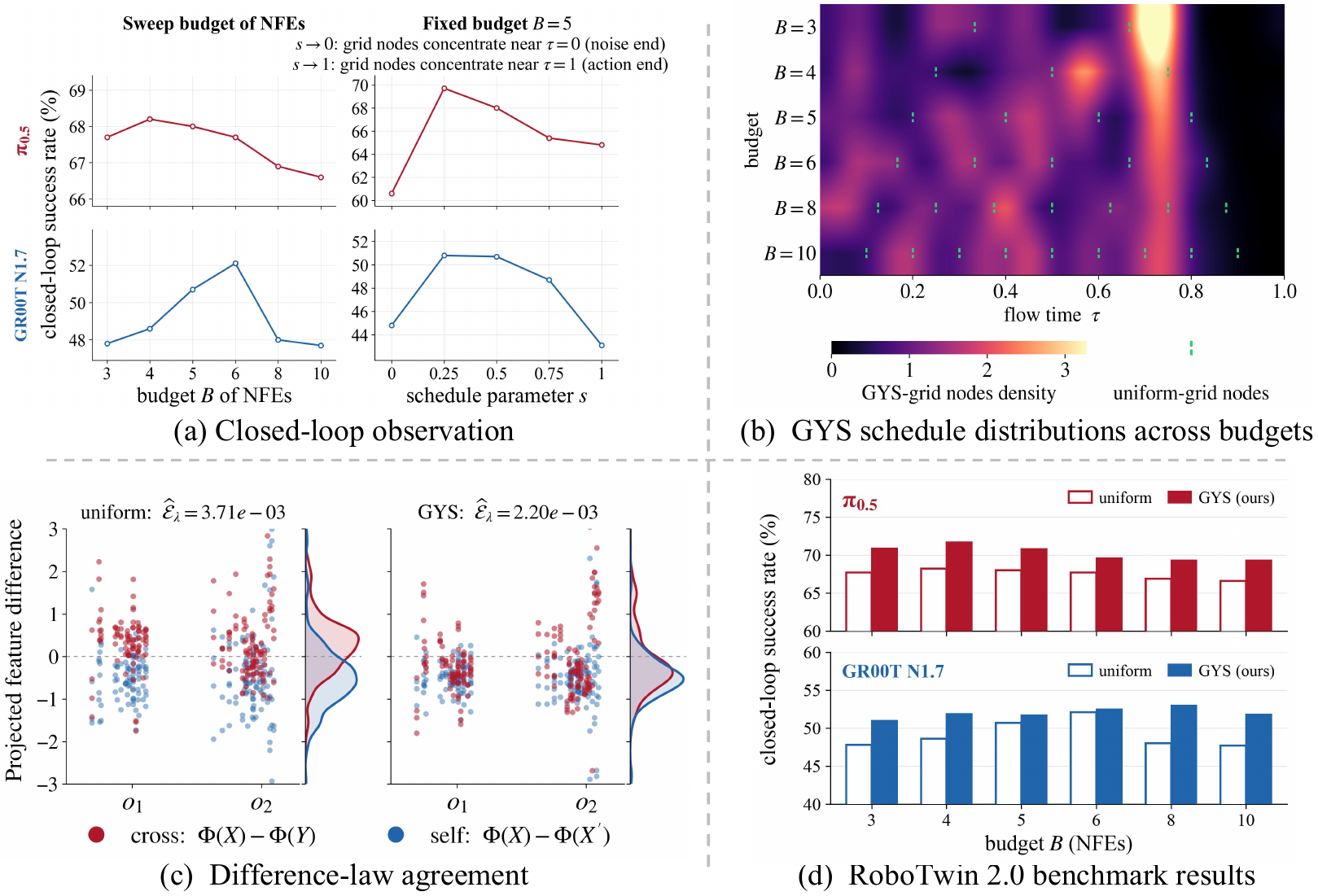}
\caption{
Motivation and effects of \method{}.
(a) Closed-loop success varies non-monotonically with the NFE budget $B$
under uniform sampling and changes with the grid at $B{=}5$
(\cref{tab:app:power-grid}), consistent with the proposed curvature and
late-time sensitivity mechanisms.
(b) Across budgets, \method{} places fewer interior sampling nodes near the
action endpoint, where deviations may be amplified; green ticks mark
uniform-grid nodes.
(c) For two observation groups $o_1$ and $o_2$ pooled from the calibration
set, \method{} lowers the energy distance between the two transformed
pairwise-difference distributions, the objective used to match transformed
generated and demonstrated action laws.
(d) Over 50 \robotwin{} tasks, \method{} improves mean closed-loop success at
nearly every fixed budget.}
\label{fig:degradation}
\end{figure}%
}

Vision--language--action (VLA) models translate visual observations and
language instructions into robot actions
\citep{ma2024survey,kawaharazuka2025vlareview}.
Flow-based VLAs couple pretrained vision--language backbones with action
experts \citep{black2024pi0}.
Recent models use flow matching
\citep{intelligence2025pi05,bjorck2025groot} to generate these actions
in chunks \citep{zhao2023act}.
During training, the action expert predicts the velocities of straight
paths from Gaussian noise to demonstrated action chunks.
Under squared loss, the optimal prediction is the conditional average
of these velocities at a given state, flow time, and observation,
called the marginal velocity field
\citep{lipman2023flow,liu2023rectified}.
Exact integration of this field transports Gaussian noise to the
demonstrated action distribution \citep{albergo2023interpolants}.
In practice, however, the policy integrates a learned approximation to the
marginal field with finite neural function evaluations (NFEs), so the
generated action distribution can deviate from the demonstrated distribution.

Two mechanisms shape the generated action distribution under this finite-step
integration.
(i)~\emph{Trajectory curvature}---averaging differently directed path
velocities can produce curved marginal trajectories, which coarse
integration steps may fail to resolve
\citep{ma2026straightflows,kornilov2024optimal}.
(ii)~\emph{Lipschitz window}---near the action endpoint, the marginal
field can become highly sensitive to the action state, allowing small
deviations from the marginal flow to be amplified
\citep{gao2024cnf,chen2025densejump} (\cref{app:lipschitz}).
Since curvature and the amplification of small deviations vary along
flow time, evaluations at different times need not influence the
endpoint distribution equally.
Choosing these evaluation times, the sampling schedule, therefore
provides a way to change the generated distribution.
Our goal is to optimize the sampling schedule to bring the generated
distribution into agreement with the demonstrations and improve task
success for each fixed budget $B$.

To see whether these distributional effects matter in closed loop, we vary
the NFE budget and sampling schedule for \pifive{} and \groot{} across
50 \robotwin{} tasks (\cref{fig:degradation}, panel~(a)).
Under uniform sampling, success peaks at $B=4$ for \pifive{} and $B=6$
for \groot{}, then falls at larger budgets.
At fixed $B=5$, a moderate bias toward the noise endpoint outperforms
the uniform grid, whereas concentrating nodes near either endpoint
lowers success.
The non-monotonic budget trend and fixed-budget schedule sensitivity are
consistent with the combined effects of trajectory curvature and
late-time amplification.
Taken with the mechanisms above, these observations identify the
sampling schedule as a lever on both the generated action distribution
and task success.
They motivate a demonstration-grounded criterion for selecting
evaluation times at each fixed budget.

We propose \methodlong{} (\method{}), a training-free method that
grounds the sampling schedule in the demonstrations, motivated by
Drifting \citep{deng2026drifting}, which trains a generator by
driving its generated distribution into agreement with the data
distribution.
\method{}'s target is that the chunks the frozen policy generates be
distributed as the demonstrated ones at each observation.
\method{} compares candidate sampling schedules using a
\emph{truncated fractional-difference operator}: it applies finite
differences of a calibrated, possibly non-integer order along each arm
joint's action sequence. This generalizes first-order differences
between consecutive actions and second-order differences between those
changes, using a fixed window of action steps.
One demonstrated chunk per observation still cannot estimate the
demonstrated distribution, so \method{} instead compares two types of
\emph{pairwise differences}: generated--demonstrated differences and
generated--generated differences, formed at the same observation using
independent noise draws. Both are transformed by the fractional-difference
operator. Under the stated identifiability assumptions, their distributions
coincide exactly when the transformed action distributions match,
so driving them into agreement realizes this target
(\cref{sec:method:objective}).
\method{} optimizes the sampling schedule offline for each task by
minimizing an energy distance between the two transformed
pairwise-difference distributions
(\cref{fig:degradation}, panel~(c)), and the optimized schedule
replaces the uniform one.
At the same budget, \method{} brings the generated distribution closer to
the demonstrated one and raises the task success rate.

We validate \method{} on a simulation benchmark and on a real robot.
On 50 \robotwin{} tasks \citep{chen2025robotwin2} at 3 to 10 NFEs,
\method{} raises the mean closed-loop success rate from 67.5\% to
70.3\% for \pifive{} and from 49.2\% to 52.0\% for \groot{}
(\cref{fig:degradation}, panel~(d)).
On three real-world manipulation tasks at 3 and 4 NFEs, it raises
the success rate from 67.7\% to 69.5\% for \pifive{} and from 63.5\%
to 66.5\% for \groot{}.
Across nearly every fixed budget, \method{} raises the task success rate, and
the gain comes from the optimized sampling schedule alone, with the
weights and the inference cost untouched.

%% file: sections/2_related.tex
\section{Related Work}
\label{sec:related}

\paragraph{Flow matching in robot policies.}
Flow matching has become an increasingly common approach to action
generation in robot policies, appearing in
generalist VLAs
\citep{black2024pi0,shukor2025smolvla,yu2026walloss05,wu2026lingbotvla} and task-specific
visuomotor policies such as FlowPolicy \citep{zhang2025flowpolicy}
and CoLA-Flow Policy \citep{wu2026colaflow}.
These flow-matching approaches generate actions by
transporting noise through learned flows.
This shared flow-matching formulation makes the distribution mismatch discussed in
\cref{sec:intro} a concern across the family of flow-matching robot
policies.
We therefore study sampling schedule optimization to answer the following
question: Where along flow time should a
flow-matching sampler spend its evaluations to reduce this distribution
mismatch and improve the task success rate?

\paragraph{Training-based correction of distribution mismatch in flow matching.}
Curvature in the marginal flow can lead to distribution mismatch when
its trajectories are approximated by finite integration steps.
Training-based methods address this source of mismatch by changing
the velocity quantity predicted by the model.
One line explicitly straightens the flow.
Rectified flow retrains on the pairs its own flow produces, and each
round leaves the flow straighter \citep{liu2023rectified}.
Finding that recursion costly, consistency flow matching straightens
within a single round, constraining the velocities along a trajectory
to a single predicted endpoint \citep{yang2024consistencyfm}.
A second line regresses the average
velocity over an interval, accounting
for displacement along curved trajectories
\citep{geng2025meanflow}.
These methods modify the learned dynamics to ease numerical integration.
We instead keep the trained policy fixed and reduce distribution mismatch
through the sampling schedule alone. To account for both trajectory curvature and
late-time sensitivity, we select the schedule by
distributional agreement with demonstrations, thereby improving the
task success rate.

\paragraph{Optimizing the sampling schedule at inference.}
Trained VLAs deploy the default uniform grid as their sampling schedule at inference
\citep{black2024pi0,bjorck2025groot}.
To address trajectory curvature, several image-generation methods optimize sampling schedules without retraining the generative model, using a finely resolved reference trajectory or an error surrogate to guide step placement. These include
LD3 \citep{tong2025ld3}, Align Your Steps \citep{sabour2024align}, and
optimized time steps \citep{xue2024accelerating} for diffusion models, and
SharpEuler \citep{gupta2026sharpeuler} for flow models. For
a frozen robot policy, however, this objective is insufficient because
small deviations from the marginal flow can be amplified within the Lipschitz window.
Most related to our work, Dense-Jump \citep{chen2025densejump} traces
the same late-time degradation to a velocity field whose Lipschitz
constant diverges near the action end.
It therefore brings a non-uniform sampling schedule to state-based locomotion and
dexterous control, stepping densely up to a jump point and then
crossing the remaining flow times in a single step.
That one schedule serves every task and requires retraining the policy
under a U-shaped time distribution.
We instead optimize a sampling schedule for each task and budget with
the policy frozen, using a demonstration-grounded energy-distance
objective for distribution matching. GYS raises the task success rate at nearly every fixed NFE
budget. However, our approach only reduces the effect of amplified
small deviations from the marginal flow without eliminating the Lipschitz window.

%% file: sections/3_background.tex
\section{Background}
\label{sec:background}

\paragraph{Conditional flow matching.}
Given an observation condition $o$, a flow matching policy
\citep{lipman2023flow} transforms Gaussian noise into an action chunk
of horizon $H$ and action dimension $d$, represented as an element of
$\R^{H \times d}$.
For a VLA policy, the observation is encoded by its vision--language
backbone to condition the action expert.
\Cref{sec:method} introduces a summary $\obsrep(o)$ of this representation
for distribution comparison.
Expert demonstrations induce a conditional distribution $\demos_o$
over such chunks.
For a demonstrated chunk $Y \sim \demos_o$ and an independent noise
sample $Z \sim \mathcal{N}(0,I)$ of the same shape, rectified flow
\citep{liu2023rectified} defines the linear interpolation
\begin{equation*}
X_\tau = (1-\tau)\,Z + \tau\,Y ,
\end{equation*}
which moves from $Z$ at $\tau=0$ to $Y$ at $\tau=1$ with constant
velocity $Y-Z$.
A neural velocity field $v_\theta(x,\tau,o)$ is trained to predict this
velocity along the interpolation by minimizing
\begin{equation}
\mathcal{L}(\theta)
= \E_{o}\,
\E_{Y \sim \demos_o,\; Z \sim \mathcal{N}(0,I),\; \tau}
\bigl\lVert v_\theta(X_\tau, \tau, o) - (Y - Z) \bigr\rVert^2 .
\label{eq:cfm}
\end{equation}
Here, $o$ is drawn from the demonstrations, while $\tau$ is drawn
independently of $(Y,Z)$ from the policy's training-time distribution
on $[0,1]$.
At a fixed $(x,\tau,o)$, different pairs $(Y,Z)$ may induce
interpolation paths that pass through the same state $x$, with
different velocities $Y-Z$.
Under squared loss, the population-optimal prediction is therefore their
conditional mean, known as the marginal velocity field,
\begin{equation}
v^\ast(x, \tau, o)
= \E\bigl[\, Y - Z \;\big|\; X_\tau = x,\; \tau,\; o \,\bigr] .
\label{eq:vstar}
\end{equation}
Equivalently, \cref{eq:cfm} differs from
$\E\lVert v_\theta-v^\ast\rVert^2$ only by a term independent of
$\theta$, so training on the sampled velocities $Y-Z$ fits the
marginal field $v^\ast$.

Under standard regularity conditions, the exact marginal flow follows
the marginal distributions of the linear interpolant
\citep{lipman2023flow,albergo2023interpolants}.
Let $X^\ast(\tau)$ denote this flow initialized at
$X^\ast(0)\sim\mathcal{N}(0,I)$. For almost every $o$,
\begin{equation*}
\Law\bigl(X^\ast(\tau)\mid o\bigr)
= \Law\bigl(X_\tau\mid o\bigr),
\qquad \tau\in[0,1).
\end{equation*}
As $\tau$ approaches $1$, its distribution therefore converges to the
demonstrated action distribution $\demos_o$.

\paragraph{Inference as numerical integration.}
At inference, the policy draws a fresh noise sample and integrates the
learned field according to
\begin{equation}
\frac{\mathrm{d}X(\tau)}{\mathrm{d}\tau}
= v_\theta\bigl(X(\tau), \tau, o\bigr),
\qquad X(0) = Z \sim \mathcal{N}(0,I),
\label{eq:ode}
\end{equation}
and executes the endpoint $X(1)$ as an action chunk.
In practice, the ODE is solved under a budget of $B$ neural function
evaluations.
We represent a sampling schedule by an integration grid
$\sched=(\tau_0,\tau_1,\ldots,\tau_B)$ with
$0=\tau_0<\tau_1<\cdots<\tau_B=1$, and let $\grids_B$ denote
the set of such schedules.
On any $\sched\in\grids_B$, explicit Euler discretizes \cref{eq:ode}
using numerical states $X(\tau_k)$ satisfying
\begin{equation}
X(\tau_k) = X(\tau_{k-1})
+ (\tau_k - \tau_{k-1})\,
v_\theta\bigl(X(\tau_{k-1}), \tau_{k-1}, o\bigr),
\qquad k = 1, \dots, B,
\label{eq:euler}
\end{equation}
thereby defining a generator
$\Psi^{\sched}_\theta(Z;o)=X(\tau_B)$ and its conditional output law
$\gen^{\sched}_o$.
Trained VLAs default to the uniform schedule $\tau_k=k/B$.\footnote{The
released \pifive{} sampler integrates from $\tau=1$ (noise) to
$\tau=0$ (action); we write every grid from $0$ (noise) to $1$
(action). By default \pifive{} uses $B=10$ and \groot{} uses
$B=4$.}
For fixed parameters $\theta$ and budget $B$, schedules in $\grids_B$
therefore define a family of generators with the same NFE budget,
$\{\Psi^{\sched}_\theta:\sched\in\grids_B\}$, which differ only in their
placement of integration steps along flow time.

%% file: sections/4_method.tex
\section{Method}
\label{sec:method}

\method{} selects the sampling schedule offline, and the three
subsections follow one chain of reasoning (\cref{fig:motivation}).
\Cref{sec:method:problem} states what the schedule should achieve, namely that the
generated action distribution should match the demonstrated one after a
transformation that makes the comparison informative.
This target involves the demonstrated distribution, which the data do not
provide directly, so \cref{sec:method:objective} builds an objective from
pairwise differences that the data can supply and that vanishes exactly
when the target holds.
\Cref{sec:method:solve} evaluates this objective on a calibration set and
searches for the schedule with the lowest score (\cref{alg:gys} in \cref{app:procedure}).

\afterpage{%
\begin{figure}[htb]
\centering
\includegraphics[width=\linewidth]{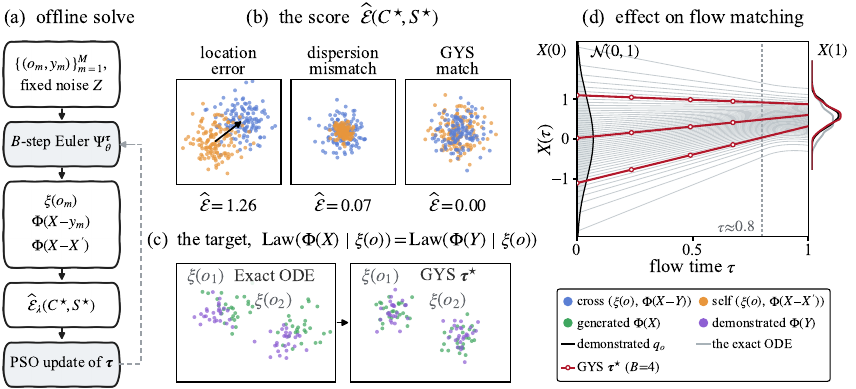}
\caption{Distributional schedule selection in \method{}.
\emph{(a)} With policy weights and the NFE budget fixed, \method{}
searches grids offline using calibration observations and shared noise
draws.
\emph{(b)} An energy score compares both types of transformed pairwise differences
paired with observation summaries, revealing location and dispersion
mismatch.
\emph{(c)} Aligning these joint laws targets conditional agreement
between transformed generated and demonstrated action distributions,
illustrated for two observation groups.
\emph{(d)} The schematic shows how stepping across a late-time sensitive
region can reduce endpoint mismatch relative to exact integration of the
learned ODE.}
\label{fig:motivation}
\end{figure}%
}

\subsection{The distribution-matching target}
\label{sec:method:problem}

Let $\{(o_m,y_m)\}_{m=1}^{M}$ be a calibration set of demonstration
pairs, where $o_m$ is an observation and $y_m$ its recorded action chunk.
The flow-matching action expert receives a representation of $o_m$
from the vision--language backbone: a prefix key--value cache in
\pifive{} and backbone features in \groot{}.
For distribution comparison, we summarize the backbone hidden states
by their mean over non-padding tokens and denote this summary by
$\obsrep(o)\in\R^{\Dh}$ (\cref{eq:obsrep}).
The policy continues to use its full conditioning representation.
For the analysis below, we assume that the generated and demonstrated
conditional laws depend on $o$ only through $\obsrep(o)$.
This is a sufficiency assumption on the summary, not a property
guaranteed by mean pooling.

For fixed policy weights $\theta$ and NFE budget $B$, each schedule
$\sched\in\grids_B$ defines a generator $\Psi^{\sched}_\theta$.
Let $\gen^{\sched}_{\obsrep}$ and $\demos_{\obsrep}$ denote the generated
and demonstrated action-chunk laws conditional on $\obsrep$.
Our ideal target is equality of these laws,
\begin{equation}
\gen^{\sched}_{\obsrep}=\demos_{\obsrep}
\qquad \text{for almost every }\obsrep .
\label{eq:target}
\end{equation}

However, directly comparing raw action chunks is sensitive to
schedule-invariant biases between the frozen policy and demonstrations.
To remove these biases, we compare chunks through a linear
\emph{truncated fractional-difference operator}
$\horizonop_{\bandord}:\R^{H\times d}\to\R^F$
\citep{hosking1981fractional,podlubny1999fractional}.
It selects the arm-joint coordinates, applies fractional differencing of
order $\bandord$ along each joint's action sequence, and vectorizes the
valid output positions. Each output uses a window of $K=10$ action steps.
For a joint sequence $u[h]$, order one gives $u[h]-u[h-1]$;
order two gives $u[h]-2u[h-1]+u[h-2]$.
Non-integer orders extend these finite differences through generalized
binomial weights. Positive orders can reduce the contribution of offsets
shared across action steps and emphasize changes within the chunk;
negative orders instead perform fractional summation.
The operation acts along the action horizon, not along flow time.
\Cref{app:horizon} gives the weights and truncation convention.

We compare schedules using the resulting transformed target,
\begin{equation}
\Law\bigl(\horizonop_{\bandord}(X)\mid\obsrep\bigr)
=\Law\bigl(\horizonop_{\bandord}(Y)\mid\obsrep\bigr)
\qquad\text{for almost every }\obsrep,
\label{eq:target-phi}
\end{equation}
where $X\sim\gen^{\sched}_{\obsrep}$ and
$Y\sim\demos_{\obsrep}$.
Equality of the full laws implies \cref{eq:target-phi}, but the converse
need not hold because the operator discards some action coordinates and
horizon information.
The differencing order $\bandord$ is calibrated
once for each task and budget so that generated and demonstrated chunks are comparable at this order, and the schedule is optimized for agreement at that order (\cref{sec:method:solve}).
\subsection{An energy-distance objective from pairwise-difference distributions}
\label{sec:method:objective}

The target in \cref{eq:target-phi} compares generated chunks with the conditional demonstrated law, but the calibration set contains only one demonstrated chunk per observation, so this law cannot be estimated directly.
We use instead pairwise differences that the data can supply.
At a fixed observation, the transformed target makes generated and
demonstrated chunks equal in distribution after the fractional-difference operator.
Consider the pairwise difference between a generated chunk and an independent
demonstration. Replacing that demonstration with another independent
generated chunk therefore leaves the transformed pairwise-difference distribution
unchanged.
We use this property to compare generated--demonstrated differences with
generated--generated differences.

Fix a schedule $\sched$, a differencing order $\bandord$, and a condition
$\obsrep$.
Let $Y\sim\demos_{\obsrep}$ and let $X,X'\sim\gen^{\sched}_{\obsrep}$
be mutually independent conditional on $\obsrep$.
We form the \emph{generated--demonstrated pairwise difference} $X-Y$
and the \emph{generated--generated pairwise difference} $X-X'$.
Because the fractional-difference operator is linear, transforming each
pairwise difference is equivalent to subtracting the transformed chunks.
Define their conditional laws by
\begin{equation}
\crosslaw_{\obsrep}
= \Law\bigl(\horizonop_{\bandord}(X - Y) \mid \obsrep\bigr),
\qquad
\selflaw_{\obsrep}
= \Law\bigl(\horizonop_{\bandord}(X - X') \mid \obsrep\bigr) .
\label{eq:pairwise-laws}
\end{equation}
The replacement argument gives
$\crosslaw_{\obsrep}=\selflaw_{\obsrep}$ under the transformed target
\cref{eq:target-phi}.

The calibration set does not provide enough demonstrations to estimate
these laws separately at every observation, so we pool samples across
the task. Different observations, however, can require different
actions. Pooling the pairwise differences alone discards their association with
the observations and can conceal differences between the conditional laws.
We therefore retain the observation summary in each sample and compare
the resulting joint laws,
\begin{equation}
\jcross = \bigl(\obsrep(o),\; \horizonop_{\bandord}(X - Y)\bigr),
\qquad
\jself = \bigl(\obsrep(o),\; \horizonop_{\bandord}(X - X')\bigr),
\label{eq:joint}
\end{equation}
where $(o,Y)$ is a demonstration pair and $X,X'$ are generated at $o$
with independent noise.
Let $C^{*}$ and $S^{*}$ denote the population laws of $\jcross$ and
$\jself$. Both use the same distribution of observations, so they
share the same $\obsrep$-marginal. Consequently,
$C^{*}=S^{*}$ is equivalent to
$\crosslaw_{\obsrep}=\selflaw_{\obsrep}$ for almost every $\obsrep$
(\cref{app:joint}).

For a joint sample $(\obsrep,\phi)\in\R^{\Dh+F}$, where $\phi$ is a
transformed pairwise difference, define the weighted Euclidean metric
\begin{equation}
\mathcal{D}_\lambda\bigl((\obsrep, \phi), (\obsrep', \phi')\bigr)
= \sqrt{\,\lambda^2\,\lVert \obsrep - \obsrep' \rVert^2
+ \lVert \phi - \phi' \rVert^2\,} ,
\label{eq:metric}
\end{equation}
where $\lambda>0$ sets the relative scale of the observation coordinates.
We use this metric to define the energy distance between $C^{*}$ and
$S^{*}$ \citep{szekely2013energy}. It compares distances between samples
from different distributions with distances between samples from the
same distribution.
For mutually independent draws
$\jcross, \tilde{\jcross} \sim C^{*}$ and
$\jself, \tilde{\jself} \sim S^{*}$,
\begin{equation}
\mathcal{E}_\lambda\bigl(C^{*}, S^{*}\bigr)
= 2\,\E\,\mathcal{D}_\lambda(\jcross, \jself)
- \E\,\mathcal{D}_\lambda(\jcross, \tilde{\jcross})
- \E\,\mathcal{D}_\lambda(\jself, \tilde{\jself}) .
\label{eq:energy}
\end{equation}

\begin{assumption}
\label{as:identifiability}
Fix a sampling schedule $\sched$ and a differencing order $\bandord$.
The conditional generated and demonstrated laws depend on $o$ only
through $\obsrep(o)$.
The joint laws $C^{*}$ and $S^{*}$ have finite first moments.
For almost every value of $\obsrep$, the characteristic function of
$\Law(\horizonop_{\bandord}(X)\mid \obsrep)$ has no open set of zeros.
\end{assumption}

The final condition permits recovery of the transformed action law from its
pairwise-difference law; it is not guaranteed by Gaussian input noise alone.

\begin{proposition}
\label{prop:identifiability}
Under \cref{as:identifiability}, for every $\lambda>0$,
\begin{equation*}
\mathcal{E}_\lambda(C^{*},S^{*})=0
\quad\Longleftrightarrow\quad
\Law\bigl(\horizonop_{\bandord}(X)\mid \obsrep\bigr)
=\Law\bigl(\horizonop_{\bandord}(Y)\mid \obsrep\bigr)
\quad\text{for almost every }\obsrep .
\end{equation*}
\end{proposition}

\Cref{app:theory} gives the proof.
The positive observation weight preserves the conditional comparison;
setting $\lambda=0$ retains only the pooled pairwise-difference marginals
(\cref{app:joint}).
The proposition concerns equality of population distributions and does
not bound distribution error or task success in terms of an empirical
score. We use the empirical score to compare sampling schedules and
evaluate the selected schedules in closed loop in \cref{sec:experiments}.

\subsection{Offline schedule optimization}
\label{sec:method:solve}

We now evaluate the objective in \cref{eq:energy} on the calibration set to score each candidate schedule and search for the schedule with the lowest score (\cref{fig:motivation}(a)).
For each task and NFE budget, \method{} optimizes the sampling schedule
using a fixed calibration set and frozen policy weights, without
environment interaction.
We parameterize the schedule by its step sizes
$h=(h_1,\dots,h_B)$, with nodes $\tau_k=\sum_{j=1}^{k}h_j$, and
search over
\begin{equation}
\feas
= \Bigl\{\, h \in \R^B \;\Bigm|\;
\textstyle\sum_{k=1}^{B} h_k = 1,\;\;
h_k \geq \varepsilon_h \,\Bigr\},
\qquad \varepsilon_h=10^{-3},
\label{eq:feasible}
\end{equation}
This leaves $B-1$ free parameters while ensuring positive step sizes.

For each observation $o_m$, draw an even number $N\geq4$ of independent
Gaussian noise samples and reuse them for every schedule.
For a candidate $\sched$, let
$X_{m,i}=\Psi^{\sched}_\theta(Z_{m,i};o_m)$ denote the resulting chunks.
Split them into two disjoint groups of size $N_c=N/2$.
The first group supplies generated--demonstrated differences against $y_m$; the second
supplies generated--generated differences through cyclic pairing.
The joint samples are
\begin{equation}
\begin{aligned}
\jcross_{m,i}
  &= \bigl(\obsrep(o_m),\;
     \horizonop_{\bandord}(X_{m,i} - y_m)\bigr),
  &\quad i &= 1, \dots, N_c , \\
\jself_{m,j}
  &= \bigl(\obsrep(o_m),\;
     \horizonop_{\bandord}(X_{m,j} - X_{m,j+1})\bigr),
  &\quad j &= N_c\!+\!1, \dots, N ,
\end{aligned}
\label{eq:samples}
\end{equation}
with cyclic index $X_{m,N+1}=X_{m,N_c+1}$.
Samples from all calibration observations are pooled, retaining the
observation summary in each sample.

We use this construction on a set of initial sampling schedules to
calibrate the comparison. We select a common differencing order $\qstar$
by balancing the spreads of the two pairwise-difference distributions on
these schedules, and set $\lambda$ by matching
the typical scales of the transformed pairwise differences and standardized observation
summaries. Both parameters and the observation standardization remain
fixed throughout the search (\cref{app:objective}).

We then score each candidate schedule by the energy distance between the
empirical joint distributions of \cref{eq:samples}, using standardized
observation coordinates in \cref{eq:metric}. Including all diagonal
terms, the score is
\begin{equation}
\begin{aligned}
\energy_\lambda(\bandord;\, \sched)
={}& \frac{2}{n_{\jcross}\, n_{\jself}}
  \sum_{m,i} \sum_{m',j}
  \mathcal{D}_\lambda\bigl(\jcross_{m,i}, \jself_{m',j}\bigr)
- \frac{1}{n_{\jcross}^{2}}
  \sum_{m,i} \sum_{m',i'}
  \mathcal{D}_\lambda\bigl(\jcross_{m,i}, \jcross_{m',i'}\bigr)
\\[2pt]
&- \frac{1}{n_{\jself}^{2}}
  \sum_{m,j} \sum_{m',j'}
  \mathcal{D}_\lambda\bigl(\jself_{m,j}, \jself_{m',j'}\bigr) ,
  \qquad n_{\jcross} = M N_c , \quad n_{\jself} = M(N - N_c) .
\end{aligned}
\label{eq:energyhat}
\end{equation}
At the calibrated order $\qstar$, we solve
\begin{equation}
\obj(h)=\energy_\lambda\bigl(\qstar;\sched(h)\bigr),
\qquad \min_{h\in\feas}\obj(h).
\label{eq:loss}
\end{equation}
Reusing the calibration pairs and noise samples makes this a
deterministic comparison of schedules under the same metric.

We search the feasible set with particle swarm optimization
\citep[PSO;][]{clerc2002particle}, which does not require differentiating
through the policy. The particles start at the initial schedules,
and each update is projected onto $\feas$.
We return the lowest-scoring schedule among all candidates evaluated,
$\sched^\star$; \cref{app:procedure} gives the full procedure and
computational cost.
At deployment, $\sched^\star$ replaces the uniform schedule at the same
NFE budget, with no additional policy evaluations.

%% file: sections/5_experiments.tex
\section{Experiments}
\label{sec:experiments}

\begin{figure}[htb]
\centering
\includegraphics[width=0.9\linewidth]{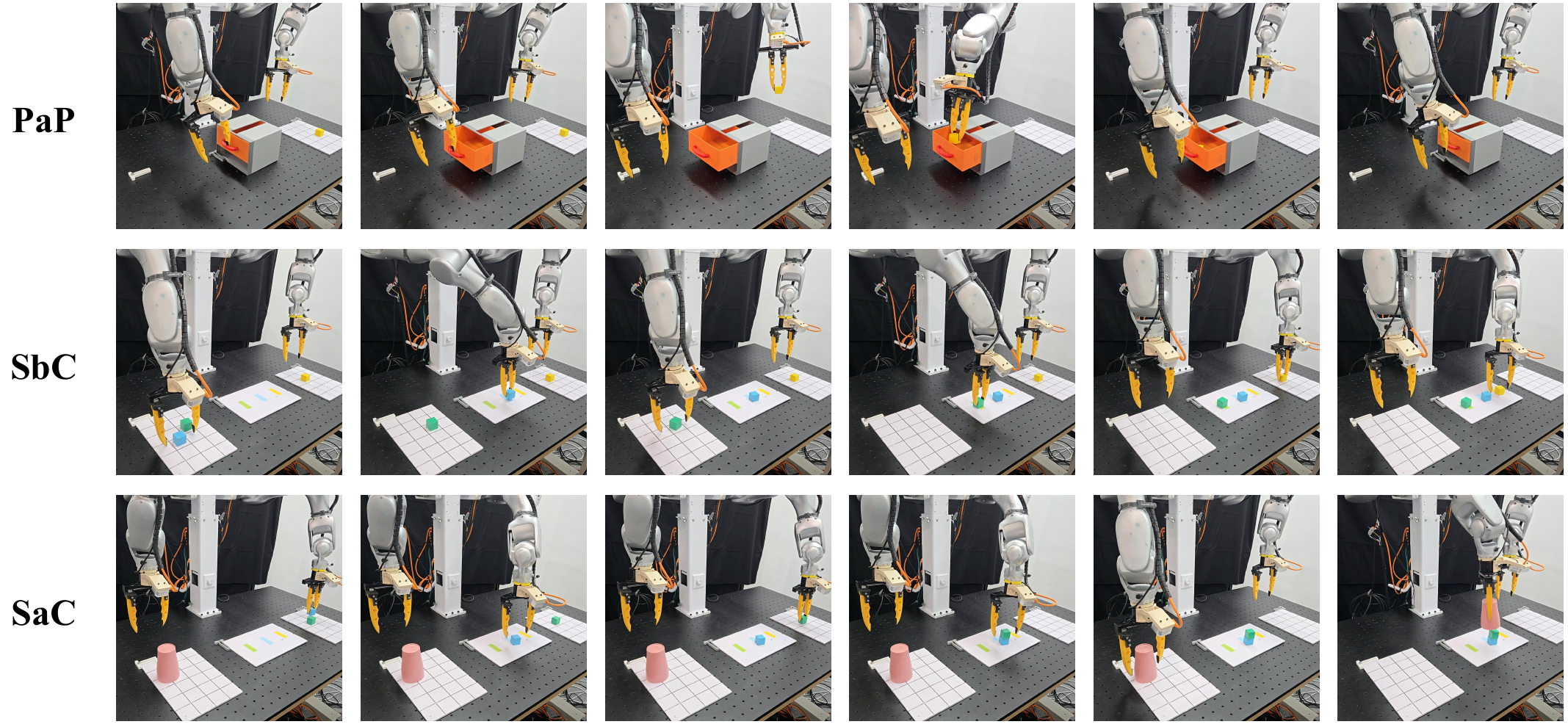}
\caption{Visualization of the three real-world manipulation tasks.
Each row shows one task as a sequence of frames ordered in time, from
the initial scene (left) to task completion (right).
The task prompts are, for \pap{}, ``Pull the drawer open and place the
yellow cube inside''; for \sbc{}, ``Place the green, blue, and yellow
cubes into the regions of matching color''; and for \sac{}, ``First
stack the blue cube and the green cube in the designated region, then
cover them with the upside-down cup''.}
\label{fig:realworld}
\end{figure}

\paragraph{Experiment setup.}
We evaluate \method{} on a simulation benchmark and on a real robot.
Simulated evaluation runs on \robotwin{} \citep{chen2025robotwin2}
with its dual-arm Aloha-AgileX embodiment and 50 manipulation tasks.
Real-world evaluation runs on a ROKAE AR5 dual-arm platform with
three tasks of increasing difficulty, pulling a drawer and picking
cubes (\pap{}), sorting cubes by color (\sbc{}), and stacking cubes
and covering them (\sac{}).
\Cref{fig:realworld} shows the three real-world tasks with their full
prompts.
We fine-tune \pifive{} and \groot{} by co-training on the official
clean datasets for \robotwin{} and on teleoperated demonstrations for
the real robot;
\Cref{app:setup} gives the collection and training
details.
Each \robotwin{} task--budget cell is evaluated for 100 episodes, and
we report the mean success rate.
Within a task every method runs under one inference configuration,
sharing the prompt, the list of scene seeds, and the sequence of
initial-noise seeds bound to each scene seed, so that corresponding
chunks are generated from the same noise as far as possible.
We compare \method{} at budgets $B \in \{3, 4, 5, 6, 8, 10\}$ against the uniform schedule, the released default
of both policies; SharpEuler, whose schedule uses finer steps where the flow
trajectory bends fastest, adapted to the action expert; and
Dense-Jump, whose schedule uses $B{-}1$ uniform Euler steps on $[0, 0.5]$
and crosses the rest in a single terminal jump.
Dense-Jump is the only method evaluated on its own checkpoints.
Dense-Jump fine-tunes each policy with flow time drawn from a U-shaped Beta
distribution instead of the policy's default\footnote{On the flow-time
axis of \cref{sec:background}, both released implementations draw the
time from $\mathrm{Beta}(1.0, 1.5)$, which concentrates its mass at
the noise end.}.
\cref{app:baselines} lists their sampling schedules.
On the real robot we compare \method{} at $B \in \{3, 4\}$ against uniform only, with gridded boards fixing the object placements so
every method runs the same 100-episode benchmark.

\paragraph{Main results.}
\Cref{tab:main} addresses our central question: can an offline schedule
selected from demonstrations improve the closed-loop behavior of a fixed
flow-matching VLA under the same inference budget? Across 50 \robotwin{}
tasks, two VLA policies, and six NFE budgets, \method{} consistently
shifts performance upward relative to the uniform default. It improves
mean success by $2.8$ percentage points for both \pifive{} and \groot{}, with gains at almost every fixed budget.
The improvement across architectures and NFE regimes shows that the
uniform discretization leaves meaningful task performance unrealized,
and that demonstration-grounded schedule selection provides a more
effective allocation of the available inference steps.

Because \method{} changes only the sampling schedule, its gains require
neither policy retraining nor additional deployment-time computation at
a fixed budget. More importantly, the optimized schedules improve the
success--latency trade-off: for \pifive{}, \method{} at $B{=}4$ and
$B{=}6$ outperforms the 10-step uniform sampler while reducing per-chunk
latency by $35\%$ and $24\%$, respectively. The same principle transfers
beyond simulation. On the real robot, \method{} improves the three-task
average for both policies and both tested budgets, with gains of
$1.7$--$3.3$ percentage points (\cref{tab:realworld}). Thus, schedule
optimization improves the deployment-time success--latency frontier of a
fixed VLA without modifying its parameters.

The comparison with alternative schedulers further clarifies what the
schedule should optimize. SharpEuler allocates steps according to the
geometry of the learned flow, whereas Dense-Jump uses a single
dense-then-jump schedule shared across tasks. Neither yields comparable
closed-loop success, and \method{} achieves the highest average success
for both policies. This comparison supports our central design choice:
for VLA action experts, the sampling schedule should be selected by how
well its outputs agree with task demonstrations, rather than solely by
ODE integration accuracy or a task-agnostic allocation of flow time.

\begin{table}[htb]
\caption{Simulation success rate (\%) at $B \in \{3, 4, 5, 6, 8, 10\}$.
Each score is averaged over the 50 \robotwin{} tasks at 100 episodes
per task.
For Uniform and \method{}, we run three additional evaluation rounds under the same configuration and report the mean and sample standard deviation across the four rounds.
The Avg.\ column is the mean over the six budgets, and its subscript is
the sample standard deviation across them.
The Gain row is \method{} minus Uniform.
Latency (ms) is the mean per-chunk inference time on one NVIDIA L4,
shared across methods at each $B$.}
\label{tab:main}
\centering
\fontsize{8}{10}\selectfont
\setlength{\tabcolsep}{5pt}
\begin{tabular}{llccccccc}
\toprule
Policy & Schedule & $B{=}3$ & $B{=}4$ & $B{=}5$ & $B{=}6$ & $B{=}8$ & $B{=}10$ & Avg. \\
\midrule
\multirow{6}{*}{\pifive{}}
 & Uniform            & 67.7\textsubscript{\(\pm\)0.6} & 68.2\textsubscript{\(\pm\)0.4} & 68.0\textsubscript{\(\pm\)0.5} & 67.7\textsubscript{\(\pm\)0.4} & 66.9\textsubscript{\(\pm\)0.4} & 66.6\textsubscript{\(\pm\)0.2} & 67.5\textsubscript{\(\pm\)0.6} \\
 & SharpEuler         & 67.5 & 67.4 & 66.8 & 66.5 & 65.0 & 64.4 & 66.3\textsubscript{\(\pm\)1.3} \\
 & Dense-Jump         & 59.1 & 60.1 & 59.6 & 59.1 & 60.1 & 59.7 & 59.6\textsubscript{\(\pm\)0.4} \\
 & \cellcolor{gray!20}\method{} (ours) & \cellcolor{gray!20}\textbf{70.9\textsubscript{\(\pm\)0.3}} & \cellcolor{gray!20}\textbf{71.7\textsubscript{\(\pm\)0.3}} & \cellcolor{gray!20}\textbf{70.8\textsubscript{\(\pm\)0.2}} & \cellcolor{gray!20}\textbf{69.6\textsubscript{\(\pm\)0.4}} & \cellcolor{gray!20}\textbf{69.3\textsubscript{\(\pm\)0.2}} & \cellcolor{gray!20}\textbf{69.3\textsubscript{\(\pm\)0.3}} & \cellcolor{gray!20}\textbf{70.3\textsubscript{\(\pm\)1.0}} \\
\cmidrule(l){2-9}
 & Gain               & +3.2 & +3.5 & +2.8 & +1.9 & +2.4 & +2.7 & +2.8 \\
 & Latency (ms)       & 185 & 196 & 215 & 230 & 269 & 302 & -- \\
\midrule
\multirow{6}{*}{\groot{}}
 & Uniform            & 47.8\textsubscript{\(\pm\)0.7} & 48.6\textsubscript{\(\pm\)0.6} & 50.7\textsubscript{\(\pm\)0.7} & 52.1\textsubscript{\(\pm\)0.5} & 48.0\textsubscript{\(\pm\)0.6} & 47.7\textsubscript{\(\pm\)0.5} & 49.2\textsubscript{\(\pm\)1.8} \\
 & SharpEuler         & 47.7 & 46.1 & 47.6 & 43.6 & 43.3 & 42.0 & 45.1\textsubscript{\(\pm\)2.4} \\
 & Dense-Jump         & 44.6 & 44.3 & 44.7 & 44.8 & 44.2 & 45.0 & 44.6\textsubscript{\(\pm\)0.3} \\
 & \cellcolor{gray!20}\method{} (ours) & \cellcolor{gray!20}\textbf{51.0\textsubscript{\(\pm\)0.5}} & \cellcolor{gray!20}\textbf{51.9\textsubscript{\(\pm\)0.4}} & \cellcolor{gray!20}\textbf{51.7\textsubscript{\(\pm\)0.4}} & \cellcolor{gray!20}\textbf{52.5\textsubscript{\(\pm\)0.6}} & \cellcolor{gray!20}\textbf{53.0\textsubscript{\(\pm\)0.5}} & \cellcolor{gray!20}\textbf{51.8\textsubscript{\(\pm\)0.5}} & \cellcolor{gray!20}\textbf{52.0\textsubscript{\(\pm\)0.7}} \\
\cmidrule(l){2-9}
 & Gain               & +3.2 & +3.3 & +1.0 & +0.4 & +5.0 & +4.1 & +2.8 \\
 & Latency (ms)       & 123 & 148 & 158 & 175 & 206 & 242 & -- \\
\bottomrule
\end{tabular}
\end{table}

\begin{figure}[htb]
\centering
\begin{minipage}[t]{0.53\linewidth}
\vspace{0pt}
\centering
\newsavebox{\realworldtablebox}
\sbox{\realworldtablebox}{%
\fontsize{8}{10}\selectfont
\setlength{\tabcolsep}{1.8pt}
\renewcommand{\arraystretch}{1.1}
\begin{tabular}{llcccccccc}
\toprule
 & & \multicolumn{4}{c}{$B{=}3$} & \multicolumn{4}{c}{$B{=}4$} \\
\cmidrule(lr){3-6} \cmidrule(lr){7-10}
Policy & Schedule & \pap{} & \sbc{} & \sac{} & Avg. & \pap{} & \sbc{} & \sac{} & Avg. \\
\midrule
\multirow[b]{2}{*}{\pifive{}}
 & Uniform          & \textbf{78} & 71 & 52 & 67.0 & 77 & 73 & 55 & 68.3 \\
 & \cellcolor{gray!20}\method{} (ours) & \cellcolor{gray!20}77 & \cellcolor{gray!20}\textbf{74} & \cellcolor{gray!20}\textbf{55} & \cellcolor{gray!20}\textbf{68.7} & \cellcolor{gray!20}\textbf{80} & \cellcolor{gray!20}\textbf{75} & \cellcolor{gray!20}\textbf{56} & \cellcolor{gray!20}\textbf{70.3} \\
\midrule
\multirow[b]{2}{*}{\shortstack[l]{GR00T\\N1.7}}
 & Uniform          & 72 & 67 & 47 & 62.0 & 75 & 69 & 51 & 65.0 \\
 & \cellcolor{gray!20}\method{} (ours) & \cellcolor{gray!20}\textbf{74} & \cellcolor{gray!20}\textbf{70} & \cellcolor{gray!20}\textbf{52} & \cellcolor{gray!20}\textbf{65.3} & \cellcolor{gray!20}75 & \cellcolor{gray!20}\textbf{73} & \cellcolor{gray!20}\textbf{55} & \cellcolor{gray!20}\textbf{67.7} \\
\bottomrule
\end{tabular}}
\captionsetup{width=\wd\realworldtablebox}
\captionof{table}{Real robot success rate (\%) at $B{=}3$\,\&\,$4$ on the shared 100-episode benchmark.}
\label{tab:realworld}
\usebox{\realworldtablebox}
\end{minipage}\hfill
\begin{minipage}[t]{0.45\linewidth}
\vspace{0pt}
\centering
\includegraphics[width=\linewidth]{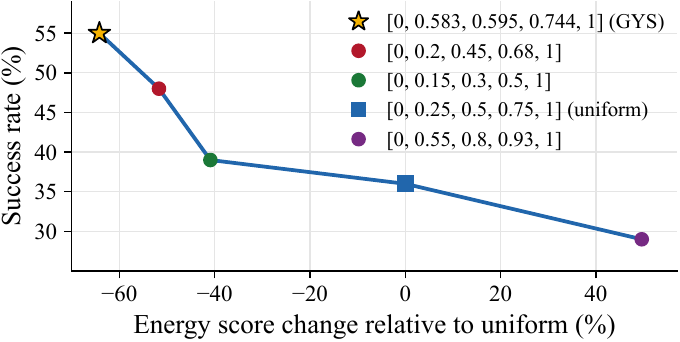}
\captionof{figure}{The relationship between energy distance and task success.}
\label{fig:energy-success}
\end{minipage}
\end{figure}

\begin{table}[htb]
\caption{Ablation and generalization success rates (\%) at $B{=}3$\,\&\,$4$ on
\robotwin{}.
The ablation Full column uses $N{=}8$ and $M{=}200$.
In w/o self, ``self'' denotes the generated--generated
pairwise difference $X-X'$.
The generalization study evaluates VLAs fine-tuned on the clean demonstrations of \robotwin{} in previously unseen random scenes (Clean2Random).\protect\footnotemark}
\label{tab:ablation}
\centering
\fontsize{8}{10}\selectfont
\setlength{\tabcolsep}{2pt}
\begin{tabular}{lc c >{\columncolor{gray!20}}c cccccccc c >{\columncolor{gray!20}}c}
\toprule
 & & & \multicolumn{9}{c}{\method{} Ablation Study} & \multicolumn{2}{c}{Generalization Study} \\
\cmidrule{4-12} \cmidrule(l){13-14}
Policy & NFEs & Uniform & Full & $\bandord{=}0$ & $\bandord{=}1$ & $\lambda{=}0$ & $N{=}4$ & $N{=}16$ & $N{=}32$ & $M{=}400$ & w/o self & \makebox[36pt]{Uniform} & \makebox[36pt]{Full} \\
\midrule
\multirow[b]{2}{*}{\pifive{}}
 & 3 & 67.7 & 70.9 & 68.1 & 68.0 & 70.2 & 71.0 & \textbf{71.4} & 70.6 & 70.8 & 63.0 & 41.2 & \textbf{44.5} \\
 & 4 & 68.2 & \textbf{71.7} & 66.1 & 67.9 & 70.5 & 70.8 & 71.2 & 70.9 & 71.5 & 62.6 & 42.7 & \textbf{45.9} \\
\midrule
\multirow[b]{2}{*}{\shortstack[l]{GR00T\\N1.7}}
 & 3 & 47.8 & 51.0 & 38.8 & 47.3 & 51.0 & \textbf{51.3} & 50.5 & 50.8 & 50.6 & 44.9 & 15.0 & \textbf{19.6} \\
 & 4 & 48.6 & 51.9 & 46.2 & 46.3 & 51.1 & 51.7 & \textbf{52.4} & 51.5 & 52.0 & 44.2 & 16.2 & \textbf{21.2} \\
\bottomrule
\end{tabular}
\end{table}
\footnotetext{Except for the generalization study, all \robotwin{} evaluations in this paper use Clean2Clean.}

\paragraph{Ablation study.}
\Cref{tab:ablation} evaluates the objective formulation, sensitivity to the optimization parameters ($\bandord, \lambda$), and sensitivity to the sample sizes $N$ and $M$. For each variant, we re-optimize the sampling schedule at $B{=}3$ and $B{=}4$. Full denotes \method{} with the defaults $N{=}8$ and $M{=}200$.
We first ablate the fractional differencing order $\bandord$.
Compared with the calibrated $\bandord$ in Full, fixing $\bandord$ to compare raw positions ($\bandord{=}0$) or first-order differences ($\bandord{=}1$) reduces success rates by $2.8$--$5.6$ percentage points on \pifive{} and by $3.7$--$12.2$ percentage points on \groot{}.
This confirms that schedule optimization requires calibrating $\bandord$ to each task and budget rather than adopting a fixed transformation order.
We next ablate the observation weight $\lambda$. Setting $\lambda{=}0$
does not diminish the empirical effectiveness of \method{}: the resulting
schedules continue to outperform uniform for both policies and budgets.
We nevertheless retain $\lambda>0$ because the
observation block preserves the conditional comparison: under the stated
assumptions, the population objective vanishes if and only if the transformed
conditional distributions match
(\cref{app:joint,prop:identifiability}).
Finally, performance is stable across $N\in\{4,8,16,32\}$, with every
variant lying within $0.9$ percentage points of Full. This robustness indicates that
the default $N{=}8$ provides an effective accuracy--cost trade-off for
the pooled, task-level objective.
Doubling the calibration set from $M{=}200$ to $M{=}400$
changes success rates by at most $0.4$ percentage points across policies
and budgets. We set $M{=}200$ by default to balance optimization
efficiency and task performance.

The \emph{w/o self} variant scores schedules using only
generated--demonstrated paired differences. With the same calibrated order
$\qstar$ as Full, it minimizes the mean transformed residual
\begin{equation*}
\obj_{\mathrm{w/o\ self}}(h)
= \frac{1}{MN}\sum_{m=1}^{M}\sum_{i=1}^{N}
  \bigl\lVert\horizonop_{\qstar}(X_{m,i}-y_m)\bigr\rVert,
\end{equation*}
where $X_{m,i}$ is generated at $o_m$ using $\sched(h)$, as in
\cref{eq:samples}. Each residual is measured against zero at its own
observation, so
$\mathcal{D}_{\lambda}((\obsrep(o_m),\phi),
(\obsrep(o_m),0))=\lVert\phi\rVert$.
This objective pulls each generated chunk toward its paired demonstration,
often crowding integration steps near the action endpoint. Such crowding can
collapse a four-step schedule to nearly two effective steps while placing
multiple steps where the \emph{Lipschitz window} can arise. For instance, with \pifive{} at $B{=}4$, \emph{w/o self} yields
$[0, 0.107, 0.964, 0.999, 1]$ on blocks ranking rgb and
$[0, 0.395, 0.987, 0.999, 1]$ on beat block hammer.
Compared with Full, w/o self reduces success by $7.9$--$9.1$ percentage points
on \pifive{} and by $6.1$--$7.7$ percentage points on \groot{}.
The performance gap shows that generated--demonstrated paired differences
alone are not enough and that accounting for generated--generated paired
differences is important for effective schedule optimization.

We investigate the relationship between energy distance and task success on
blocks ranking rgb, a \robotwin{} task, with \groot{} at $B{=}4$.
\Cref{fig:energy-success} compares five sampling schedules using the same
$\lambda$ and $\nu$ selected in the original offline optimization. Across these
schedules, lower energy scores correspond to higher success rates. The
\method{} schedule reduces the energy score by $64.2\%$ relative to uniform
and raises the success rate from $36\%$ to $55\%$.

\paragraph{Generalization study.}
We evaluate whether sampling schedules optimized on clean demonstrations
generalize to unseen domain-randomized environments. Using the VLAs fine-tuned
on the clean dataset and the per-task schedules optimized on the clean calibration
set without re-optimization, we test closed-loop performance across the 50 \robotwin{} tasks
in unseen random scenes. In \cref{tab:ablation}, across $B{=}3$ and $B{=}4$, \method{}
improves mean success over the uniform schedule by $3.3$ percentage points on \pifive{}
and by $4.8$ percentage points on \groot{}. These gains show that for frozen general-purpose
VLAs, schedules optimized offline from limited demonstrations generalize across visual
and spatial variations to improve task success rate in randomized scenes.

%% file: sections/6_conclusion.tex
\section{Conclusion and Limitations}
\label{sec:conclusion}
\label{sec:limitations}

We discuss two mechanisms affecting the generated distribution in
flow-matching policies, trajectory curvature and the Lipschitz window,
and introduce \methodlong{} (\method{}), a training-free method for
optimizing sampling schedules.
\method{} directly targets agreement between the generated and
demonstrated distributions through a fractional-difference operator.
Experiments on \robotwin{} and a real robot with \pifive{} and
\groot{} show that the optimized schedules improve mean task success
at nearly every fixed NFE budget, without additional
inference overhead.

\paragraph{Limitations.}
Although \method{} improves distributional agreement after truncated
fractional differencing along the action horizon and raises task success,
it does not eliminate either mechanism because the learned
velocity field remains unchanged. This is the trade-off of a
training-free design, as \method{} mitigates their combined
effects by optimizing the sampling schedule.

%% file: sections/appendix.tex
\section{Theoretical analysis}
\label{app:theory}

Fix the sampling schedule $\sched$ and the linear fractional-difference operator
$\horizonop_{\bandord}$. Under \cref{as:identifiability}, the population
claim in \cref{prop:identifiability} follows from three equivalences
for every fixed $\lambda>0$:
\begin{equation}
\begin{aligned}
&\Law\bigl(\horizonop_{\bandord}(X)\mid \obsrep\bigr)
=\Law\bigl(\horizonop_{\bandord}(Y)\mid \obsrep\bigr)
\\[-1pt]
&\quad\Longleftrightarrow\quad
\crosslaw_{\obsrep}=\selflaw_{\obsrep}
\quad\Longleftrightarrow\quad
C^{*}=S^{*}
\quad\Longleftrightarrow\quad
\mathcal{E}_\lambda(C^{*},S^{*})=0 ,
\end{aligned}
\label{eq:chain}
\end{equation}
The first equivalence holds for almost every condition and is proved by
conditional deconvolution in \cref{app:deconv}. The second uses the
common condition marginal of the two joint laws
(\cref{app:joint}). The last follows from the characterization of
equality by Euclidean energy distance for any fixed $\lambda>0$
\citep{szekely2013energy}. These equivalences characterize when the
energy-distance objective vanishes.

\subsection{Deconvolution of the pairwise-difference laws}
\label{app:deconv}

We prove the first link of \cref{eq:chain}: given $\obsrep$,
\begin{equation*}
\Law\bigl(\horizonop_{\bandord}(X)\mid\obsrep\bigr)
=\Law\bigl(\horizonop_{\bandord}(Y)\mid\obsrep\bigr)
\quad\Longleftrightarrow\quad
\Law\bigl(\horizonop_{\bandord}(X-Y)\mid\obsrep\bigr)
=\Law\bigl(\horizonop_{\bandord}(X-X')\mid\obsrep\bigr).
\end{equation*}
All statements are conditional on $\obsrep$, in the sense of a regular
conditional distribution on $\R^{\Dh}$, and hold for almost every value
of $\obsrep$ under \cref{as:identifiability}.

\emph{Setup.} Generate $X,X'$ at observation $o$ from independent noise
draws, also independent of $(o,Y)$. Since their conditional laws and
the demonstration law depend on $o$ only through $\obsrep(o)$ under
\cref{as:identifiability}, $X,X',Y$ remain mutually independent given
$\obsrep$, with $X$ and $X'$ identically distributed. This independence
is used in the characteristic-function factorizations below. For any
action-space random element $W$, write
\begin{equation*}
\varphi_W(\omega)
=\E\bigl[\exp(i\langle\omega,\horizonop_{\bandord}(W)\rangle)
\mid\obsrep\bigr],
\qquad\omega\in\R^{F},
\end{equation*}
for the conditional characteristic function of its transformed value.

\emph{Necessity.} Suppose
$\Law(\horizonop_{\bandord}(X)\mid\obsrep)
=\Law(\horizonop_{\bandord}(Y)\mid\obsrep)$.
Then $\horizonop_{\bandord}(Y)$ and $\horizonop_{\bandord}(X')$
share a law given $\obsrep$ and are both independent of
$\horizonop_{\bandord}(X)$. By linearity,
\begin{equation*}
\begin{aligned}
\horizonop_{\bandord}(X-Y)
&=\horizonop_{\bandord}(X)-\horizonop_{\bandord}(Y),\\
\horizonop_{\bandord}(X-X')
&=\horizonop_{\bandord}(X)-\horizonop_{\bandord}(X').
\end{aligned}
\end{equation*}
The two transformed pairwise differences therefore share a conditional law:
$\crosslaw_{\obsrep}=\selflaw_{\obsrep}$.

\emph{Sufficiency.} Suppose
$\crosslaw_{\obsrep}=\selflaw_{\obsrep}$. Conditional independence
and linearity of the fractional-difference operator give
\begin{align}
\varphi_{X-Y}(\omega)
&=\E\bigl[\exp(i\langle\omega,\horizonop_{\bandord}(X-Y)\rangle)
\mid\obsrep\bigr]
\nonumber\\
&=\varphi_X(\omega)\,\overline{\varphi_Y(\omega)}.
\label{eq:app:cf-cross}
\end{align}
The same factorization with $X'$ in place of $Y$ yields
\begin{equation}
\varphi_{X-X'}(\omega)
=\varphi_X(\omega)\,\overline{\varphi_X(\omega)}
=\bigl\lvert\varphi_X(\omega)\bigr\rvert^2.
\label{eq:app:cf-self}
\end{equation}
The assumed equality of pairwise-difference laws now gives
$\varphi_X(\omega)\,\overline{\varphi_Y(\omega)}
=\bigl\lvert\varphi_X(\omega)\bigr\rvert^2$ for every $\omega$, or
equivalently
\begin{equation}
\varphi_X(\omega)
\Bigl(\overline{\varphi_Y(\omega)}-\overline{\varphi_X(\omega)}\Bigr)
=0
\qquad\text{for every }\omega.
\label{eq:app:cf-factor}
\end{equation}
Under \cref{as:identifiability}, the zero set
$\{\omega:\varphi_X(\omega)=0\}$ has empty interior. Its complement
is dense, and division by $\varphi_X$ in \cref{eq:app:cf-factor} gives
equality there. Continuity then implies
$\varphi_X(\omega)=\varphi_Y(\omega)$ for every $\omega$.
A characteristic function determines its law, so
$\Law(\horizonop_{\bandord}(X)\mid\obsrep)
=\Law(\horizonop_{\bandord}(Y)\mid\obsrep)$.

\subsection{From conditional laws to the energy-distance objective}
\label{app:joint}

We prove the second link of \cref{eq:chain}: 
\begin{equation*}
\begin{aligned}
&\Law\bigl(\horizonop_{\bandord}(X-Y)\mid\obsrep\bigr)
=\Law\bigl(\horizonop_{\bandord}(X-X')\mid\obsrep\bigr)
\qquad\text{for almost every } \obsrep
\\
&\quad\Longleftrightarrow\quad
\Law\bigl(\obsrep(o),\,\horizonop_{\bandord}(X-Y)\bigr)
=\Law\bigl(\obsrep(o),\,\horizonop_{\bandord}(X-X')\bigr).
\end{aligned}
\end{equation*}

\emph{Setup.} By \cref{eq:joint}, $C^{*}$ and $S^{*}$ are the laws of the pairs
$\jcross = (\obsrep(o), \horizonop_{\bandord}(X-Y))$ and
$\jself = (\obsrep(o), \horizonop_{\bandord}(X-X'))$ on
$\R^{\Dh} \times \R^{F}$.
Both joint laws have the same $\obsrep$ marginal
$\mu=\Law(\obsrep(o))$.
For almost every $\obsrep$, their conditional pairwise-difference laws are
$\crosslaw_{\obsrep}$ and $\selflaw_{\obsrep}$, respectively, by
\cref{eq:pairwise-laws}.
Since the marginal is shared, disintegration writes each joint law as
\begin{equation*}
\begin{aligned}
C^{*}(A \times B) &= \int_{A} \crosslaw_{\obsrep}(B) \, \mu(d\obsrep),\\
S^{*}(A \times B) &= \int_{A} \selflaw_{\obsrep}(B) \, \mu(d\obsrep),
\end{aligned}
\end{equation*}
for measurable $A\subseteq\R^{\Dh}$ and $B\subseteq\R^{F}$.

\emph{Necessity.} If $\crosslaw_{\obsrep} = \selflaw_{\obsrep}$ for
almost every $\obsrep$, the two integrals agree for every $A$ and $B$.
In particular, the joint cumulative distribution functions coincide,
so $C^{*}=S^{*}$.

\emph{Sufficiency.} If $C^{*}=S^{*}$, the same identities give, for every
measurable $A$ and $B$,
\begin{equation*}
\int_A\bigl[\crosslaw_{\obsrep}(B)-\selflaw_{\obsrep}(B)\bigr]
\,\mu(d\obsrep)=0.
\end{equation*}
For each $B$, equality for all $A$ implies equality of the two
conditional probabilities for $\mu$-almost every $\obsrep$.
Taking $B$ in a countable determining class for $\R^F$ gives one
common null set; uniqueness of probability measures then yields
$\crosslaw_{\obsrep}=\selflaw_{\obsrep}$ for $\mu$-almost every $\obsrep$.

For $\lambda>0$, define the invertible linear map
$T_\lambda(\xi,\phi)=(\lambda\xi,\phi)$. Then
$\mathcal{D}_\lambda(w,w')=\lVert T_\lambda w-T_\lambda w'\rVert$.
The finite-first-moment assumption is preserved by $T_\lambda$, so
Euclidean energy distance is nonnegative and vanishes exactly when
the transformed joint laws agree \citep{szekely2013energy}.
Invertibility of $T_\lambda$ makes this equivalent to $C^{*}=S^{*}$,
completing the third link of \cref{eq:chain}.

At $\lambda=0$, the metric ignores $\obsrep$. A zero objective then
identifies only the pooled laws of
$\horizonop_{\bandord}(X-Y)$ and
$\horizonop_{\bandord}(X-X')$, not their conditional laws given
$\obsrep$.

\subsection{The Lipschitz window of the marginal field}
\label{app:lipschitz}

We derive the spatial Jacobian of the exact marginal field and identify
when it has an expanding direction. We use \emph{Lipschitz window} for
a late-time region in which small deviations from the marginal flow
can be amplified. A large spatial Lipschitz constant alone does not
imply amplification: the sign of the Jacobian's largest eigenvalue
matters.

\emph{Setup.} Fix an observation $o$ and identify each action chunk with
a vector in $\R^{Hd}$. Assume $\E[\lVert Y\rVert^2\mid o]<\infty$.
A demonstrated chunk $Y\sim\demos_o$ is paired with an independent noise sample
$Z \sim \mathcal{N}(0,I)$ of the same shape, and the straight
interpolant
\begin{equation}
X_\tau = (1-\tau)\,Z + \tau\,Y ,
\qquad \tau \in [0,1],
\label{eq:app:lipschitz-interp}
\end{equation}
runs from the noise at $\tau = 0$ to the chunk at $\tau = 1$ with
constant velocity $Y - Z$.
The population minimizer of the conditional flow matching loss is the
marginal velocity field
\begin{equation}
v^\ast(x,\tau,o) = \E\bigl[\,Y - Z \mid X_\tau = x,\, o\,\bigr],
\label{eq:app:lipschitz-vstar}
\end{equation}
the conditional average velocity of the interpolants that pass through
state $x$ at time $\tau$.
A perfectly trained policy attains this field and generates a chunk by
integrating it from a fresh noise sample,
\begin{equation}
\frac{\mathrm{d}X(\tau)}{\mathrm{d}\tau}
= v^\ast\bigl(X(\tau),\tau,o\bigr),
\qquad X(0) = Z \sim \mathcal{N}(0,I).
\label{eq:app:lipschitz-ode}
\end{equation}
The calculations below concern $0<\tau<1$; statements about
$\tau=1$ are limits, since the endpoint field can be singular.

\emph{The field and its Jacobian.} Along the interpolant
\eqref{eq:app:lipschitz-interp}, the noise is
recovered from the state as $Z = (X_\tau - \tau Y)/(1-\tau)$, so the
interpolant velocity $Y - Z$ that the field regresses onto becomes
\begin{equation*}
Y - Z = \frac{Y - X_\tau}{1-\tau},
\end{equation*}
and taking the conditional expectation in \eqref{eq:app:lipschitz-vstar}
writes the marginal field as a rescaled posterior mean,
\begin{equation}
v^\ast(x,\tau,o) = \frac{m(x,\tau,o) - x}{1-\tau},
\qquad
m(x,\tau,o) = \E\bigl[\,Y \mid X_\tau = x,\, o\,\bigr] .
\label{eq:app:lipschitz-field}
\end{equation}
For $0<\tau<1$, conditioning on $Y=y$ gives the Gaussian likelihood
\begin{equation*}
X_\tau \mid Y = y
\;\sim\;
\mathcal{N}\bigl(\tau y,\, (1-\tau)^2 I\bigr),
\end{equation*}
with density
\begin{equation}
p(x \mid y,\tau)
=
C(\tau)\,
\exp\!\Bigl(-\tfrac{\lVert x - \tau y\rVert^2}{2(1-\tau)^2}\Bigr),
\label{eq:app:lipschitz-lik-dens}
\end{equation}
where $C(\tau)$ is independent of $y$. Bayes' rule weights the
demonstration law by this likelihood, whether $\demos_o$ has a density
or is discrete. For fixed $0<\tau<1$, the Gaussian factor makes the
posterior moments finite and permits differentiation under the integral,
locally uniformly in $x$. The logarithmic derivative of the normalized
posterior weight with respect to $x$ is
$\tau(y-m)/(1-\tau)^2$. Hence, with
$\Sigma(x,\tau,o)=\Cov(Y\mid X_\tau=x,o)$,
\begin{equation}
\frac{\partial m}{\partial x}
=
\frac{\tau}{(1-\tau)^2}\,\Sigma(x,\tau,o).
\label{eq:app:lipschitz-dm-dx}
\end{equation}

Combining \eqref{eq:app:lipschitz-field} and
\eqref{eq:app:lipschitz-dm-dx} gives the spatial Jacobian
\citep{gao2024cnf},
\begin{equation}
J(x,\tau,o):=\frac{\partial v^\ast}{\partial x}(x,\tau,o)
= \frac{\tau}{(1-\tau)^3}\,\Sigma(x,\tau,o)
\;-\; \frac{1}{1-\tau}\,I .
\label{eq:app:lipschitz-jac}
\end{equation}
Since $J$ is symmetric, a locally expanding direction exists exactly when
\begin{equation}
\lambda_{\max}(J)>0
\quad\Longleftrightarrow\quad
\tau\,\lambda_{\max}(\Sigma(x,\tau,o))>(1-\tau)^2.
\label{eq:app:lipschitz-expansion}
\end{equation}
For multimodal demonstration laws, uncertainty between modes can
keep the posterior covariance large near mode boundaries. Wherever
\cref{eq:app:lipschitz-expansion} holds, the marginal field has a
locally expanding direction.

\section{Implementation details}
\label{app:implementation}

\subsection{Truncated fractional differencing along the action horizon}
\label{app:horizon}

For an action chunk $X\in\R^{H\times d}$, let
$u\in\R^{H\times d_a}$ contain its arm-joint coordinates after
omitting the two gripper columns, with $d_a=12$.
The operator in \cref{sec:method:problem} applies the truncated
binomial expansion of the backward-difference operator to each joint
\citep{hosking1981fractional,podlubny1999fractional}:
\begin{equation}
\horizonop_{\bandord}(X)
= \operatorname{vec}\bigl(\Delta^{\bandord}_K u\bigr),
\qquad
(\Delta^{\bandord}_K u)[h]
= \sum_{k=0}^{K-1} (-1)^k \binom{\bandord}{k}\,u[h-k].
\label{eq:horizonmap}
\end{equation}
Here $\binom{\bandord}{0}=1$ and
$\binom{\bandord}{k}=\prod_{j=0}^{k-1}(\bandord-j)/k!$ for $k\geq1$.
We use $K=10$ and retain only the valid positions $h=K,\ldots,H$,
so the output dimension is $F=(H-K+1)d_a$.
At $\bandord=0$, the operator returns the unfiltered arm-joint values
on these same positions. Thus the zero-order ablation changes the
differencing operation while preserving the horizon crop.
Negative orders use the same finite sum and correspond to truncated
fractional summation. For non-integer orders, the truncated coefficients
need not sum to zero, so constant offsets need not be removed exactly.
The order is calibrated as described next.

\subsection{Implementation of the objective}
\label{app:objective}

The empirical objective in \cref{sec:method:solve} requires observation
summaries, a common differencing order, and an observation weight.
We compute these quantities from the calibration set and initial
candidate schedules, then hold them fixed during optimization.

\paragraph{Observation summary.}
Let $\mathcal{V}(o)$ be the set of non-padding token positions and
$h_t^{(L)}(o)\in\R^{\Dh}$ the final-layer backbone state at position $t$.
The observation summary is
\begin{equation}
\obsrep(o)
= \frac{1}{\lvert\mathcal{V}(o)\rvert}
  \sum_{t\in\mathcal{V}(o)} h_t^{(L)}(o).
\label{eq:obsrep}
\end{equation}
These are the prefix states in \pifive{} and the backbone features
attended to by the action head in \groot{}.
We standardize each coordinate across the $M$ calibration observations,
using the resulting means and scales for every candidate schedule.
Write the standardized summaries as $\tilde{\obsrep}_m$.
The policy itself continues to use its full conditioning representation.

\paragraph{Differencing order.}
Let $h_p^{(0)}$, $p=1,\ldots,P$, be the initial candidate step-size
vectors specified in \cref{app:procedure}.
For each candidate, construct the pairwise differences as in
\cref{eq:samples}, pooling them across calibration observations.
At order $\bandord$, denote their transformed coordinates by
$\phi^{\jcross}_{p,i}(\bandord)$ and
$\phi^{\jself}_{p,j}(\bandord)$, with respective sample counts
$n_\jcross$ and $n_\jself$.
To calibrate the relative spread of the two types of differences,
compute the difference in their mean pairwise distances:
\begin{equation}
\varepsilon_p(\bandord)
= \frac{1}{n_\jcross^{2}}
 \sum_{i=1}^{n_\jcross}\sum_{i'=1}^{n_\jcross}
 \bigl\lVert\phi^{\jcross}_{p,i}(\bandord)
            -\phi^{\jcross}_{p,i'}(\bandord)\bigr\rVert
-\frac{1}{n_\jself^{2}}
 \sum_{j=1}^{n_\jself}\sum_{j'=1}^{n_\jself}
 \bigl\lVert\phi^{\jself}_{p,j}(\bandord)
            -\phi^{\jself}_{p,j'}(\bandord)\bigr\rVert.
\label{eq:eps}
\end{equation}
For each $p$, evaluate $\varepsilon_p$ at 121 equally spaced orders
in $[-1,2]$, in increasing order.
If a grid value is exactly zero, choose the first such order as $r_p$.
Otherwise, obtain $r_p$ by linear interpolation across the first pair
of adjacent grid values with opposite signs.
If neither case occurs, calibration fails and the search is not run.
The common order is the arithmetic mean
$\qstar=P^{-1}\sum_{p=1}^{P}r_p$.
This aggregation defines one comparison operator for the search;
it need not make $\varepsilon_p(\qstar)$ zero for every candidate.

\paragraph{Observation weight.}
Apply the operator at the common order $\qstar$ to the cached chunks
from all $P$ initial candidates. Let $\{\phi_a\}$ pool both types of
transformed pairwise differences across these candidates and observations.
We match the scales of the two coordinate blocks by setting
\begin{equation}
\lambda
= \frac{\operatorname{median}_{a\ne b}
        \lVert\phi_a-\phi_b\rVert}
       {\operatorname{median}_{m\ne m'}
        \lVert\tilde{\obsrep}_m-\tilde{\obsrep}_{m'}\rVert}.
\label{eq:lambda}
\end{equation}
The joint samples used in \cref{eq:energyhat} contain these standardized
observation coordinates. The order, weight, and standardization
statistics remain fixed for all subsequent objective evaluations.

\subsection{Offline schedule optimization and its cost}
\label{app:procedure}

\paragraph{Initialization.}
We initialize $P=24$ candidate schedules with the uniform schedule,
four fixed nonuniform schedules, and $P-5$ random schedules.
The four fixed schedules have nodes $\tau_k=g_\kappa(k/B)$, where
$g_\kappa(s)=(e^{\kappa s}-1)/(e^\kappa-1)$ and
$\kappa\in\{-2,-1,1,2\}$.
Positive $\kappa$ concentrates nodes near the noise endpoint, and
negative $\kappa$ concentrates them near the action endpoint.
For each random schedule, we draw its step-size vector from
$\operatorname{Dirichlet}(1,\ldots,1)$.
We project the initial step-size vectors onto $\feas$.
All candidate evaluations use the same $MN$ independent Gaussian
noise draws and the same calibration observations; the random seed is $0$.
\Cref{alg:gys} gives the calibration and search procedure.

\begin{algorithm}[H]
\caption{\method{} schedule optimization (one task, budget $B$)}
\label{alg:gys}
\begin{algorithmic}[1]
\Require calibration pairs $\{(o_m, y_m)\}_{m=1}^{M}$; frozen field
$v_\theta$; NFE budget $B$; seed count $P$; PSO iterations $R$
\State draw and fix $N$ noise samples $\{z_{m,i}\}$ for every $o_m$
\State $\obsrep(o_m) \gets$ mean of backbone hidden states over
non-padding tokens \Comment{\cref{eq:obsrep}}
\State $\mathcal{S} \gets \{\text{uniform schedule},\ \text{four geometric anchors},\
P{-}5\ \text{Dirichlet schedules}\}$ projected onto $\feas$
\Comment{seed round}
\For{each seed schedule $\sched \in \mathcal{S}$}
  \State generate chunks via \cref{eq:euler}; form both types of pairwise differences
  \Comment{\cref{eq:samples}}
  \State $r_{\sched} \gets$ root of the mean pairwise-distance gap \Comment{\cref{eq:eps}}
\EndFor
\State $\qstar \gets \operatorname{mean}_{\sched} r_{\sched}$;\quad
$\lambda \gets$ median heuristic at $\qstar$
\Comment{\cref{eq:lambda}}
\State $\obj(\sched) \gets \energy_\lambda(\qstar; \sched)$ for all
$\sched \in \mathcal{S}$ \Comment{empirical score, \cref{eq:loss}}
\State place one particle at each seed schedule; run PSO for $R$ iterations,
projecting each particle onto $\feas$ \Comment{\cref{app:pso}}
\State \Return $\sched^{\star} = \arg\min \obj$ over all schedules evaluated
\end{algorithmic}
\end{algorithm}

\paragraph{Particle swarm.}
\label{app:pso}
Each initial candidate supplies one particle.
We use Clerc--Kennedy constriction with $\chi=0.7298$ and
$c_1=c_2=2.05$, and a ring neighborhood
\citep{clerc2002particle}.
Each update enforces the minimum-step constraints in $\feas$ and
places the sampling nodes on a $10^{-3}$ lattice.
The order, observation weight, and noise draws are not updated.
We run $R=3$ iterations and return the lowest-scoring candidate among
all initial and subsequent evaluations.
Thus the search evaluates $P(1+R)=96$ candidates, counting repeated
schedules. \Cref{tab:convergence} reports the empirical objective reductions.

\begin{table}[htb]
\centering
\caption{Empirical objective reductions (\%) on \groot{}, averaged
across 50 \robotwin{} tasks. All reductions are normalized by the
uniform-schedule objective.}
\label{tab:convergence}
\normalsize
\begin{tabular}{lcccccc}
\toprule
Budget $B$ & 3 & 4 & 5 & 6 & 8 & 10 \\
\midrule
Initial candidates (\%) & 30.0 & 57.6 & 16.9 & 18.3 & 55.7 & 33.4 \\
Additional PSO (\%) & 9.9 & 3.2 & 12.3 & 5.9 & 1.4 & 3.1 \\
Total (\%) & 39.9 & 60.7 & 29.1 & 24.3 & 57.1 & 36.5 \\
\bottomrule
\end{tabular}
\end{table}

\paragraph{Cost and deployment.}
\label{app:cost}
We use $M=200$ calibration pairs and $N=8$ noise samples per pair.
Evaluating a previously unseen schedule generates $MN$ chunks at $B$
NFEs each. The 96 candidate evaluations therefore require at most
$96MNB$ action-expert evaluations, counted per chunk.
Repeated schedules reuse cached chunks, and calibration of $\qstar$
and $\lambda$ uses the initial candidates' cached chunks without
additional policy evaluations.

Because the calibration observations are fixed, the vision--language
backbone runs once per observation. We cache its prefix key--value
states in \pifive{} and its conditioning features in \groot{}, so
subsequent flow steps evaluate only the action expert.
The search runs on a single GPU with a micro-batch of 512 on an
A100-80\,GB or 32 on an L4-23\,GB. A micro-batch counts the
observation--noise pairs processed in one forward pass.
Candidate evaluations and their cached chunks use the same fixed
inference configuration.

\Cref{tab:cost} reports the measured offline optimization time for
\pifive{}. One Euler step over all $MN$ chunks takes $1.5$ seconds on
the A100 and $5.6$ seconds on the L4. The integration cost grows
linearly with $B$. Model loading, observation encoding, and parameter
calibration account for $8$ and $11$ minutes on the A100 and L4,
respectively.

\begin{table}[htb]
\centering
\caption{Offline optimization time for \pifive{}, in minutes for one
task and NFE budget on a single GPU.}
\label{tab:cost}
\normalsize
\begin{tabular}{llcccccc}
\toprule
GPU & Micro-batch & \multicolumn{6}{c}{Budget $B$} \\
\cmidrule(lr){3-8}
 & & 3 & 4 & 5 & 6 & 8 & 10 \\
\midrule
A100-80\,GB & 512 & 15 & 18 & 20 & 23 & 28 & 32 \\
L4-23\,GB & 32 & 38 & 47 & 56 & 65 & 83 & 101 \\
\bottomrule
\end{tabular}
\end{table}

At deployment, only the sampling schedule changes. The policy weights
and the NFE budget remain fixed, and no calibration or schedule
optimization is performed online.

\section{Experimental details}
\label{app:experiments}

\subsection{Schedule family for Figure~1(a)}
\label{app:schedule-sweep}

At $B=5$, \cref{fig:degradation}(a) compares five schedules from
the one-parameter family
\begin{equation}
\tau_k = (k/B)^{\gamma(s)},
\qquad
\gamma(s) = 3^{\,1-2s},
\qquad k=0,\ldots,B,
\label{eq:app:power-grid}
\end{equation}
where $s\in\{0,\tfrac14,\tfrac12,\tfrac34,1\}$.
The choice $s=\tfrac12$ gives the uniform grid; $s=0$ places more nodes
near the noise endpoint, while $s=1$ places more near the action endpoint.

\begin{table}[htb]
\caption{The five sampling grids at $B=5$ in \cref{fig:degradation}(a).}
\label{tab:app:power-grid}
\centering
\begin{tabular}{ccl}
\toprule
$s$ & $\gamma(s)$ & Nodes $\tau_0,\ldots,\tau_5$ \\
\midrule
$0$            & $3$     & $0,\ 0.008,\ 0.064,\ 0.216,\ 0.512,\ 1$ \\
$1/4$          & $1.732$ & $0,\ 0.062,\ 0.205,\ 0.413,\ 0.679,\ 1$ \\
$1/2$          & $1$     & $0,\ 0.200,\ 0.400,\ 0.600,\ 0.800,\ 1$ \\
$3/4$          & $0.577$ & $0,\ 0.395,\ 0.589,\ 0.745,\ 0.879,\ 1$ \\
$1$            & $0.333$ & $0,\ 0.585,\ 0.737,\ 0.843,\ 0.928,\ 1$ \\
\bottomrule
\end{tabular}
\end{table}

\subsection{Setup details}
\label{app:setup}

\paragraph{\robotwin{} configuration.}
Both policies are fine-tuned on the official \robotwin{} clean
datasets,
50 demonstration episodes for each of the 50 tasks, on the dual-arm
Aloha-AgileX embodiment.
Observations comprise a head camera and two wrist cameras, and
actions live in the 14-dimensional absolute joint space of the two
arms.

\begin{figure}[htb]
\centering
\includegraphics[width=0.6\linewidth]{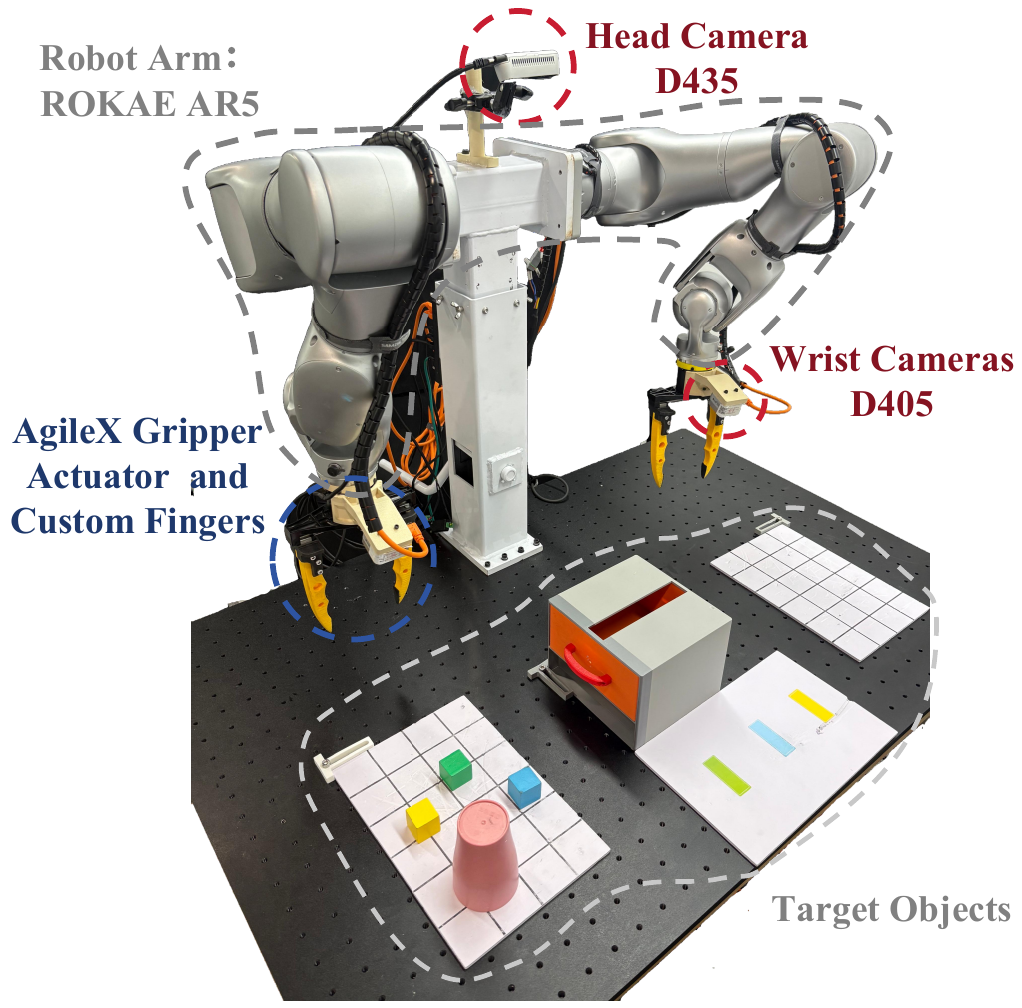}
\caption{Real-world setup.
The platform is a ROKAE AR5 dual-arm system.
A RealSense D435 supplies the head view, and a RealSense D405 on each
wrist supplies the left and right wrist views.
Each arm ends in an AgileX gripper actuator with custom fingers of
our own design.
The tabletop holds target objects the three real-world tasks
manipulate.}
\label{fig:setup}
\end{figure}

\paragraph{Real-world configuration.}
Real-world demonstrations are collected on the ROKAE AR5 dual-arm
platform of \cref{fig:setup}, whose three RGB cameras stream at
30\,Hz.
An operator teleoperates the two arms through a VR headset and hand
controllers, and every episode is recorded under the single fixed
prompt shown for its task in \cref{fig:realworld}.
In \sbc{}, the three cubes start at arbitrary positions on the two
boards, and the operator picks and places them in an unconstrained
order, keeping the number of demonstrations balanced across
the six possible orderings.
In \sac{}, the blue cube is placed in a grid cell in the upper half of
the left board and the green cube in a grid cell in the lower half,
while the cup is placed at an arbitrary position on the right board.
We record 200 episodes per task, and the resulting 600 episodes form
the co-training set.

\paragraph{Policy configuration.}
\Cref{tab:policy-config} lists the training configuration of the two
policies; all settings outside the table follow the official
releases of \pifive{} and \groot{}.

\begin{table}[htb]
\caption{Training configuration of the two policies, both fine-tuned
on $8 \times$ A800 GPUs.}
\label{tab:policy-config}
\centering
\normalsize
\begin{tabular}{lcc}
\toprule
 & \pifive{} & \groot{} \\
\midrule
Image input          & $3 \times 224 \times 224$ & $3 \times 256 \times 256$ \\
Action space         & \multicolumn{2}{c}{absolute joint (14-D \robotwin{}, 16-D real robot)} \\
Chunk length $H$     & 32 & 32 \\
Fine-tuned modules   & all & vision tower, projector, action head \\
Batch size           & 64 (8 per GPU) & 64 (8 per GPU) \\
Training steps       & 40{,}000 & 40{,}000 \\
\bottomrule
\end{tabular}
\end{table}

\subsection{Baseline definitions and adaptations}
\label{app:baselines}

The uniform schedule uses time nodes $\tau_k = k/B$.
SharpEuler optimizes the sampling schedule using the curvature of the
flow trajectory; we apply its rule to each policy's field.
Its schedule optimization is training-free and closed-form, requiring
no iterative search and using a dense reference ODE trajectory.
For each of $S = 50$ calibration chunks per task we integrate the policy's
learned field once on a dense uniform grid of $D = 100$ Euler steps at
a single fixed noise draw, and record the velocity $v_\theta$ at every
node, trimmed to the scored $(H, d)$ frame.
SharpEuler steps small where the field turns fast, so it places nodes
by the field's \emph{material acceleration} along that trajectory.
On the midpoint support $t_k = (\tau_k + \tau_{k+2})/2$ we take the
centered finite difference
\begin{equation}
\kappa(t_k)
= \Bigl\lVert
\frac{v_{\theta}(\tau_{k+1}) - v_{\theta}(\tau_k)}{(\tau_{k+2}-\tau_k)/2}
\Bigr\rVert_W ,
\label{eq:sharp-kappa}
\end{equation}
where $\lVert\cdot\rVert_W$ is the root mean square over the scored
$(H, d)$ frame, the same norm the rest of the method scores with.
The per-chunk $\kappa$ are aggregated episode-balanced---averaged
within an episode, then across episodes with equal weight---onto the
shared support.
The aggregate sets a placement density
\begin{equation}
\rho(t) \propto \bigl(\kappa(t) + \varepsilon\bigr)^{\gamma} ,
\qquad \gamma = \tfrac{1}{2} ,
\label{eq:sharp-density}
\end{equation}
which we smooth with a Gaussian kernel of standard deviation
$\sigma = 1$ grid spacing and reflection padding, then floor at
$\varepsilon = 10^{-12}$.
We extend $\rho$ to both endpoints using its boundary values and
obtain the time nodes by evaluating its inverse CDF at $B{+}1$
equally spaced quantiles on $[0,1]$, with $\tau_0=0$ and $\tau_B=1$.
Both policies use these settings and $S=50$ calibration chunks per task.

Dense-Jump uses $B{-}1$ uniform Euler steps on $[0,0.5]$ and one
terminal step to $1$. Its nodes are $\tau_k=k/(2(B{-}1))$ for
$k=0,\ldots,B{-}1$ and $\tau_B=1$.
We re-fine-tune each policy with flow time drawn from
$\mathrm{Beta}(0.5,0.5)$; all other training settings follow
\cref{app:setup}.
\Cref{tab:baseline-grids} lists the sampling schedules at each evaluated budget.

\begin{table}[htb]
\caption{Sampling schedules at every evaluated budget, written as the full node
sequence $\tau_0, \dots, \tau_B$.
Uniform and Dense-Jump are task-independent; SharpEuler and \method{}
are shown on two representative \robotwin{} tasks,
\textit{blocks\_ranking\_rgb} and \textit{stack\_bowls\_three}.
The percentage after each \method{} schedule is its empirical score
reduction over uniform on that task, $1 - L/L_{\mathrm{uniform}}$.}
\label{tab:baseline-grids}
\centering
\scriptsize
\begin{tabular}{cl b{0.32\linewidth} b{0.32\linewidth}}
\toprule
$B$ & Method & \textit{blocks\_ranking\_rgb} & \textit{stack\_bowls\_three} \\
\midrule
\multirow{6}{*}{3}
 & Uniform                 & \multicolumn{2}{l}{0, 0.333, 0.667, 1} \\
 & Dense-Jump              & \multicolumn{2}{l}{0, 0.250, 0.500, 1} \\
 & SharpEuler (\pifive{})  & 0, 0.442, 0.802, 1 & 0, 0.441, 0.800, 1 \\
 & SharpEuler (\groot{})   & 0, 0.450, 0.802, 1 & 0, 0.450, 0.802, 1 \\
 & \cellcolor{gray!20}\method{} (\pifive{})   & \cellcolor{gray!20}0, 0.27, 0.723, 1 (+20.8\%) & \cellcolor{gray!20}0, 0.175, 0.685, 1 (+13.3\%) \\
 & \cellcolor{gray!20}\method{} (\groot{})    & \cellcolor{gray!20}0, 0.218, 0.692, 1 (+38.0\%) & \cellcolor{gray!20}0, 0.475, 0.740, 1 (+47.8\%) \\
\midrule
\multirow{6}{*}{4}
 & Uniform                 & \multicolumn{2}{l}{0, 0.250, 0.500, 0.750, 1} \\
 & Dense-Jump              & \multicolumn{2}{l}{0, 0.167, 0.333, 0.500, 1} \\
 & SharpEuler (\pifive{})  & 0, 0.337, 0.632, 0.874, 1 & 0, 0.336, 0.630, 0.873, 1 \\
 & SharpEuler (\groot{})   & 0, 0.349, 0.637, 0.870, 1 & 0, 0.349, 0.637, 0.870, 1 \\
 & \cellcolor{gray!20}\method{} (\pifive{})   & \cellcolor{gray!20}0, 0.251, 0.413, 0.691, 1 (+16.2\%) & \cellcolor{gray!20}0, 0.306, 0.533, 0.774, 1 (+23.8\%) \\
 & \cellcolor{gray!20}\method{} (\groot{})    & \cellcolor{gray!20}0, 0.583, 0.595, 0.744, 1 (+64.2\%) & \cellcolor{gray!20}0, 0.071, 0.554, 0.740, 1 (+56.7\%) \\
\midrule
\multirow{6}{*}{5}
 & Uniform                 & \multicolumn{2}{l}{0, 0.200, 0.400, 0.600, 0.800, 1} \\
 & Dense-Jump              & \multicolumn{2}{l}{0, 0.125, 0.250, 0.375, 0.500, 1} \\
 & SharpEuler (\pifive{})  & 0, 0.271, 0.518, 0.738, 0.910, 1 & 0, 0.270, 0.517, 0.736, 0.909, 1 \\
 & SharpEuler (\groot{})   & 0, 0.282, 0.527, 0.740, 0.903, 1 & 0, 0.281, 0.527, 0.740, 0.903, 1 \\
 & \cellcolor{gray!20}\method{} (\pifive{})   & \cellcolor{gray!20}0, 0.233, 0.466, 0.658, 0.659, 1 (+22.8\%) & \cellcolor{gray!20}0, 0.227, 0.461, 0.545, 0.688, 1  (+28.9\%) \\
 & \cellcolor{gray!20}\method{} (\groot{})    & \cellcolor{gray!20}0, 0.298, 0.390, 0.583, 0.752, 1 (+17.0\%) & \cellcolor{gray!20}0, 0.267, 0.450, 0.451, 0.740, 1 (+25.6\%) \\
\midrule
\multirow{6}{*}{6}
 & Uniform                 & \multicolumn{2}{l}{0, 0.167, 0.333, 0.500, 0.667, 0.833, 1} \\
 & Dense-Jump              & \multicolumn{2}{l}{0, 0.100, 0.200, 0.300, 0.400, 0.500, 1} \\
 & SharpEuler (\pifive{})  & 0, 0.227, 0.442, 0.632, 0.802, 0.930, 1 & 0, 0.226, 0.441, 0.630, 0.800, 0.930, 1 \\
 & SharpEuler (\groot{})   & 0, 0.237, 0.450, 0.637, 0.802, 0.924, 1 & 0, 0.236, 0.450, 0.637, 0.802, 0.924, 1 \\
 & \cellcolor{gray!20}\method{} (\pifive{})   & \cellcolor{gray!20}0, 0.302, 0.303, 0.622, 0.623, 0.999, 1 (+25.2\%) & \cellcolor{gray!20}0, 0.317, 0.399, 0.504, 0.611, 0.863, 1 (+14.9\%) \\
 & \cellcolor{gray!20}\method{} (\groot{})    & \cellcolor{gray!20}0, 0.082, 0.162, 0.311, 0.465, 0.712, 1 (+15.2\%) & \cellcolor{gray!20}0, 0.083, 0.156, 0.300, 0.460, 0.704, 1 (+23.7\%) \\
\midrule
\multirow{6}{*}{8}
 & Uniform                 & \multicolumn{2}{l}{0, 0.125, 0.250, 0.375, 0.500, 0.625, 0.750, 0.875, 1} \\
 & Dense-Jump              & \multicolumn{2}{l}{0, 0.071, 0.143, 0.214, 0.286, 0.357, 0.429, 0.500, 1} \\
 & SharpEuler (\pifive{})  & 0, 0.170, 0.337, 0.490, 0.632, 0.763, 0.874, 0.953, 1 & 0, 0.170, 0.336, 0.489, 0.630, 0.761, 0.873, 0.953, 1 \\
 & SharpEuler (\groot{})   & 0, 0.178, 0.349, 0.499, 0.637, 0.763, 0.870, 0.949, 1 & 0, 0.177, 0.349, 0.498, 0.637, 0.763, 0.870, 0.949, 1 \\
 & \cellcolor{gray!20}\method{} (\pifive{})   & \cellcolor{gray!20}0, 0.001, 0.002, 0.194, 0.404, 0.405, 0.667, 0.668, 1 (+28.5\%) & \cellcolor{gray!20}0, 0.001, 0.24, 0.373, 0.494, 0.495, 0.644, 0.812, 1 (+16.9\%) \\
 & \cellcolor{gray!20}\method{} (\groot{})    & \cellcolor{gray!20}0, 0.037, 0.069, 0.212, 0.398, 0.399, 0.648, 0.752, 1 (+57.6\%) & \cellcolor{gray!20}0, 0.037, 0.069, 0.212, 0.398, 0.399, 0.648, 0.752, 1 (+55.2\%) \\
\midrule
\multirow{6}{*}{10}
 & Uniform                 & \multicolumn{2}{l}{0, 0.100, 0.200, 0.300, 0.400, 0.500, 0.600, 0.700, 0.800, 0.900, 1} \\
 & Dense-Jump              & \multicolumn{2}{l}{0, 0.056, 0.111, 0.167, 0.222, 0.278, 0.333, 0.389, 0.444, 0.500, 1} \\
 & SharpEuler (\pifive{})  & 0, 0.136, 0.271, 0.401, 0.518, 0.632, 0.738, 0.832, 0.910, 0.965, 1 & 0, 0.136, 0.270, 0.400, 0.517, 0.630, 0.736, 0.830, 0.909, 0.965, 1 \\
 & SharpEuler (\groot{})   & 0, 0.143, 0.282, 0.410, 0.527, 0.637, 0.740, 0.831, 0.903, 0.962, 1 & 0, 0.142, 0.281, 0.409, 0.527, 0.637, 0.740, 0.831, 0.903, 0.962, 1 \\
 & \cellcolor{gray!20}\method{} (\pifive{})   & \cellcolor{gray!20}0, 0.462, 0.546, 0.547, 0.548, 0.549, 0.550, 0.562, 0.606, 0.667, 1 (+32.8\%) & \cellcolor{gray!20}0, 0.441, 0.442, 0.443, 0.444, 0.445, 0.446, 0.596, 0.597, 0.612, 1 (+42.3\%) \\
 & \cellcolor{gray!20}\method{} (\groot{})    & \cellcolor{gray!20}0, 0.160, 0.196, 0.355, 0.444, 0.445, 0.584, 0.705, 0.751, 0.752, 1 (+25.5\%) & \cellcolor{gray!20}0, 0.160, 0.196, 0.355, 0.444, 0.445, 0.584, 0.705, 0.751, 0.752, 1 (+34.8\%) \\
\bottomrule
\end{tabular}
\end{table}

\clearpage

\subsection{Full results on the \robotwin{} benchmark}
\label{app:tables}

\begin{scriptsize}
\setlength{\tabcolsep}{5pt}
\begin{longtable}{l cccc cccc}
\caption{Per-task success rate (\%) from a single evaluation round on the 50 \robotwin{} tasks, under
the uniform schedule and the optimized \method{} schedule. \groot{} is
reported at $B = 4$ and $B = 8$, \pifive{} at $B = 5$ and $B = 10$. Each \method{} entry reports the success
rate, with the differencing order $\nu^\star$ used to optimize its schedule in
parentheses.}
\label{tab:per-task} \\
\toprule
 & \multicolumn{4}{c}{\groot{}} & \multicolumn{4}{c}{\pifive{}} \\
\cmidrule(lr){2-5} \cmidrule(lr){6-9}
 & \multicolumn{2}{c}{$B{=}4$} & \multicolumn{2}{c}{$B{=}8$}
 & \multicolumn{2}{c}{$B{=}5$} & \multicolumn{2}{c}{$B{=}10$} \\
\cmidrule(lr){2-3} \cmidrule(lr){4-5} \cmidrule(lr){6-7} \cmidrule(lr){8-9}
Task & Uniform & \method{} ($\nu^\star$) & Uniform & \method{} ($\nu^\star$)
     & Uniform & \method{} ($\nu^\star$) & Uniform & \method{} ($\nu^\star$) \\
\midrule
\endfirsthead
\toprule
Task & Uniform & \method{} ($\nu^\star$) & Uniform & \method{} ($\nu^\star$)
     & Uniform & \method{} ($\nu^\star$) & Uniform & \method{} ($\nu^\star$) \\
\midrule
\endhead
\bottomrule
\endfoot
\textit{adjust bottle} & 95 & \textbf{99}\,(0.425) & \textbf{98} & 96\,(0.45) & 100 & 100\,(1) & 100 & 100\,(1.125) \\
\textit{beat block hammer} & \textbf{67} & 64\,(0.15) & 73 & \textbf{75}\,(0.225) & 44 & \textbf{52}\,(1) & \textbf{58} & 57\,(1.15) \\
\textit{blocks ranking rgb} & 36 & \textbf{55}\,(0.4) & 41 & \textbf{52}\,(0.45) & 79 & \textbf{88}\,(1) & 81 & \textbf{88}\,(1.175) \\
\textit{blocks ranking size} & 10 & \textbf{11}\,(0.475) & \textbf{12} & 8\,(0.55) & 40 & \textbf{44}\,(0.975) & 40 & \textbf{44}\,(1.125) \\
\textit{click alarmclock} & 100 & 100\,(0.175) & 98 & \textbf{100}\,(0.275) & 98 & \textbf{100}\,(0.975) & 99 & \textbf{100}\,(1.175) \\
\textit{click bell} & \textbf{100} & 98\,(-0.05) & 97 & \textbf{98}\,(0.05) & 98 & \textbf{100}\,(0.65) & 98 & \textbf{100}\,(0.725) \\
\textit{dump bin bigbin} & 76 & \textbf{86}\,(0.4) & \textbf{93} & 86\,(0.45) & 90 & \textbf{98}\,(0.9) & \textbf{95} & 89\,(1.025) \\
\textit{grab roller} & \textbf{100} & 97\,(0.25) & \textbf{99} & 97\,(0.325) & 100 & 100\,(0.85) & 100 & 100\,(1.075) \\
\textit{handover block} & \textbf{43} & 40\,(0.45) & 38 & \textbf{51}\,(0.525) & \textbf{43} & 38\,(1) & 29 & \textbf{31}\,(1.125) \\
\textit{handover mic} & 64 & \textbf{84}\,(0.45) & \textbf{87} & 78\,(0.525) & 97 & \textbf{100}\,(1.125) & 99 & \textbf{100}\,(1.325) \\
\textit{hanging mug} & 8 & \textbf{17}\,(0.4) & 7 & \textbf{11}\,(0.45) & 8 & \textbf{12}\,(1) & 5 & \textbf{9}\,(1.15) \\
\textit{lift pot} & \textbf{66} & 55\,(0.275) & 46 & \textbf{60}\,(0.35) & \textbf{71} & 64\,(0.8) & 57 & \textbf{66}\,(0.925) \\
\textit{move can pot} & 25 & \textbf{34}\,(0.375) & 24 & \textbf{36}\,(0.475) & 53 & \textbf{56}\,(0.925) & 56 & 56\,(1.05) \\
\textit{move pillbottle pad} & 29 & \textbf{34}\,(0.35) & 26 & \textbf{44}\,(0.425) & 69 & \textbf{74}\,(0.975) & 67 & \textbf{70}\,(1.1) \\
\textit{move playingcard away} & \textbf{79} & 71\,(0.225) & \textbf{76} & 69\,(0.3) & 90 & \textbf{93}\,(1.025) & 92 & \textbf{95}\,(1.2) \\
\textit{move stapler pad} & 1 & \textbf{4}\,(0.375) & \textbf{4} & 2\,(0.45) & 15 & \textbf{25}\,(0.925) & 13 & \textbf{18}\,(1.05) \\
\textit{open laptop} & 77 & \textbf{80}\,(0.575) & \textbf{81} & 78\,(0.625) & 86 & \textbf{88}\,(0.9) & 80 & \textbf{83}\,(1) \\
\textit{open microwave} & 19 & \textbf{29}\,(0.45) & 7 & \textbf{33}\,(0.525) & 2 & \textbf{6}\,(0.85) & 1 & \textbf{4}\,(1) \\
\textit{pick diverse bottles} & 22 & \textbf{30}\,(0.35) & \textbf{34} & 33\,(0.35) & 63 & \textbf{64}\,(1.1) & 59 & \textbf{65}\,(1.225) \\
\textit{pick dual bottles} & \textbf{49} & 43\,(0.325) & 47 & \textbf{54}\,(0.375) & 82 & 82\,(1.025) & 73 & \textbf{78}\,(1.15) \\
\textit{place a2b left} & \textbf{31} & 29\,(0.35) & 31 & \textbf{32}\,(0.4) & \textbf{85} & 79\,(1) & 78 & \textbf{81}\,(1.175) \\
\textit{place a2b right} & \textbf{22} & 21\,(0.325) & 21 & \textbf{22}\,(0.375) & 75 & \textbf{82}\,(1) & 70 & \textbf{76}\,(1.225) \\
\textit{place bread basket} & \textbf{51} & 50\,(0.375) & 49 & \textbf{53}\,(0.45) & 58 & \textbf{61}\,(1.075) & 54 & \textbf{57}\,(1.275) \\
\textit{place bread skillet} & 52 & \textbf{56}\,(0.425) & 47 & \textbf{55}\,(0.5) & 76 & \textbf{79}\,(1.125) & 76 & \textbf{79}\,(1.325) \\
\textit{place burger fries} & \textbf{81} & 70\,(0.275) & 51 & \textbf{70}\,(0.325) & 89 & \textbf{92}\,(1.075) & \textbf{94} & 93\,(1.3) \\
\textit{place can basket} & 41 & \textbf{58}\,(0.425) & 43 & \textbf{59}\,(0.475) & 41 & \textbf{50}\,(0.9) & 46 & \textbf{49}\,(1.075) \\
\textit{place cans plasticbox} & \textbf{58} & 54\,(0.35) & 52 & \textbf{62}\,(0.4) & \textbf{88} & 84\,(1.025) & \textbf{72} & 68\,(1.175) \\
\textit{place container plate} & 90 & \textbf{91}\,(0.375) & \textbf{91} & 90\,(0.45) & 93 & \textbf{96}\,(1.1) & 94 & \textbf{97}\,(1.3) \\
\textit{place dual shoes} & \textbf{19} & 7\,(0.375) & \textbf{13} & 10\,(0.4) & 51 & \textbf{54}\,(0.95) & 47 & \textbf{59}\,(1.075) \\
\textit{place empty cup} & \textbf{74} & 73\,(0.225) & 73 & \textbf{79}\,(0.275) & 83 & \textbf{89}\,(0.975) & 82 & \textbf{86}\,(1.15) \\
\textit{place fan} & 18 & \textbf{23}\,(0.325) & 15 & \textbf{22}\,(0.4) & 62 & \textbf{65}\,(0.875) & 58 & \textbf{61}\,(1.025) \\
\textit{place mouse pad} & 0 & \textbf{3}\,(0.3) & \textbf{5} & 1\,(0.375) & 38 & \textbf{41}\,(0.95) & 32 & \textbf{35}\,(1.175) \\
\textit{place object basket} & 65 & \textbf{75}\,(0.45) & \textbf{72} & 61\,(0.5) & 70 & 70\,(0.975) & 66 & \textbf{73}\,(1.125) \\
\textit{place object scale} & 34 & 34\,(0.325) & \textbf{30} & 28\,(0.375) & 72 & \textbf{77}\,(0.975) & 68 & \textbf{73}\,(1.15) \\
\textit{place object stand} & 66 & \textbf{75}\,(0.3) & 61 & \textbf{77}\,(0.375) & 82 & \textbf{85}\,(1.075) & \textbf{87} & 83\,(1.3) \\
\textit{place phone stand} & \textbf{39} & 36\,(0.25) & 33 & \textbf{35}\,(0.25) & 62 & \textbf{65}\,(1.025) & \textbf{56} & 52\,(1.275) \\
\textit{place shoe} & 35 & \textbf{38}\,(0.375) & 34 & \textbf{39}\,(0.45) & 77 & \textbf{80}\,(0.875) & \textbf{80} & 79\,(1.075) \\
\textit{press stapler} & \textbf{93} & 91\,(0.45) & \textbf{93} & 91\,(0.525) & 82 & \textbf{85}\,(0.7) & 80 & \textbf{83}\,(0.75) \\
\textit{put bottles dustbin} & 16 & \textbf{42}\,(0.5) & 16 & \textbf{30}\,(0.55) & \textbf{71} & 66\,(0.9) & 68 & \textbf{70}\,(1.025) \\
\textit{put object cabinet} & 28 & 28\,(0.5) & 21 & \textbf{28}\,(0.6) & 36 & \textbf{39}\,(1.225) & 33 & \textbf{36}\,(1.475) \\
\textit{rotate qrcode} & 20 & \textbf{28}\,(0.425) & 21 & \textbf{25}\,(0.475) & 79 & \textbf{82}\,(0.95) & 71 & \textbf{78}\,(1.15) \\
\textit{scan object} & 32 & \textbf{34}\,(0.55) & 23 & \textbf{32}\,(0.6) & 17 & \textbf{21}\,(1.125) & 12 & \textbf{19}\,(1.225) \\
\textit{shake bottle} & 95 & \textbf{98}\,(0.525) & \textbf{98} & 97\,(0.6) & 100 & 100\,(1) & 100 & 100\,(1.1) \\
\textit{shake bottle horizontally} & 96 & \textbf{97}\,(0.5) & 95 & \textbf{98}\,(0.55) & 100 & 100\,(1.05) & 100 & 100\,(1.15) \\
\textit{stack blocks three} & 10 & \textbf{20}\,(0.425) & 10 & \textbf{19}\,(0.5) & 54 & \textbf{57}\,(0.875) & 42 & \textbf{45}\,(1.025) \\
\textit{stack blocks two} & 51 & \textbf{64}\,(0.35) & 44 & \textbf{67}\,(0.4) & \textbf{95} & 91\,(0.875) & 90 & \textbf{93}\,(1) \\
\textit{stack bowls three} & 38 & \textbf{52}\,(0.4) & 38 & \textbf{57}\,(0.425) & 72 & \textbf{75}\,(1.075) & 70 & \textbf{76}\,(1.225) \\
\textit{stack bowls two} & \textbf{86} & 82\,(0.35) & 75 & \textbf{87}\,(0.425) & 91 & \textbf{94}\,(0.975) & 95 & \textbf{97}\,(1.125) \\
\textit{stamp seal} & 14 & 14\,(0.325) & 12 & \textbf{21}\,(0.425) & 56 & \textbf{59}\,(1.05) & \textbf{55} & 54\,(1.2) \\
\textit{turn switch} & 25 & \textbf{34}\,(0.35) & \textbf{38} & 36\,(0.375) & \textbf{44} & 43\,(0.675) & 42 & \textbf{45}\,(0.725) \\
\midrule
\textit{Mean} & 48.9 & \textbf{52.2}\,(0.366) & 47.9 & \textbf{52.9}\,(0.428) & 68.5 & \textbf{70.9}\,(0.969) & 66.4 & \textbf{69.0}\,(1.125) \\
\end{longtable}
\end{scriptsize}